\documentclass{article}
\pdfoutput=1 
\PassOptionsToPackage{numbers, compress}{natbib}
\usepackage[preprint]{neurips_2026}

\usepackage[utf8]{inputenc}
\usepackage{hyperref}
\hypersetup{hypertexnames=false} 
\usepackage{url}
\usepackage{booktabs}
\usepackage{multirow}
\usepackage{amsfonts}
\usepackage{amsmath}
\usepackage{amssymb}
\usepackage{amsthm}
\usepackage[table]{xcolor}
\usepackage{graphicx}
\usepackage{wrapfig}
\usepackage{tikz}
\usetikzlibrary{positioning, arrows.meta, fit, calc, backgrounds, shapes.geometric}
\usepackage{algorithm}
\usepackage{algpseudocode}
\usepackage{float}

\newtheorem{proposition}{Proposition}

\title{The Platonic Defense: Backdoor Defense for Self-Supervised Encoders in the Era of Large Scale Pre-training}

\author{%
  Tuo Chen\textsuperscript{1,2} \quad
  Minjing Dong\textsuperscript{3} \quad
  Benlei Cui\textsuperscript{4} \quad
  Jian Liu\textsuperscript{2}\thanks{Corresponding authors.} \quad
  Jie Gui\textsuperscript{1,5}\footnotemark[1] \\[2pt]
  \textsuperscript{1}Southeast University \quad
  \textsuperscript{2}Ant Group \quad
  \textsuperscript{3}City University of Hong Kong \\[2pt]
  \textsuperscript{4}Alibaba Group \quad
  \textsuperscript{5}Purple Mountain Laboratories \\[2pt]
  \texttt{guijie@seu.edu.cn}
}

\begin{document}

\maketitle

\begin{abstract}
Self-supervised learning (SSL) pretrained models have become a dominant paradigm for visual representation learning, but they are vulnerable to backdoor attacks. Existing defenses struggle to defend against such attacks in a fully black-box setting because they often require access to labels, attack patterns, or training data.
To tackle this issue, we propose a new attack-agnostic, model-agnostic, and modality-agnostic black-box test-time defense paradigm, called \emph{Platonic Representation Defense}. It is inspired by the Platonic Representation Hypothesis, which suggests that large-scale independently trained encoders converge toward compatible projections of the same underlying reality. We formalize this idea as a conditional energy function defined over source representations and a set of reference representations. The energy function is trained for detection through noise-contrastive estimation and for representation purification through denoising score matching. Theoretically, the energy gap between matched and mismatched samples is lower bounded by the mutual information between source and reference representations.
We demonstrate the effectiveness of our method on multiple self-supervised encoders and more than 10 attacks. The method can perform both representation detection and purification, and achieves substantial performance gains across multiple attacks. Code is available \href{https://github.com/jsrdcht/Platonic-Representation-Defense}{here}.
\end{abstract}

\section{Introduction}
\label{sec:intro}

Artificial intelligence systems are rapidly moving toward the era of large-scale pretraining. Earlier computer vision systems often rely on task-specific solutions for image classification~\cite{he2016deep}, semantic segmentation~\cite{zhou2017scene} and other tasks. Modern large models can handle all these tasks with a single set of weights~\cite{radford2021Learning,openai2024gpt4technicalreport}, which are built on general SSL encoders that inherit strong representation ability and basic understanding of the world from pretraining~\cite{liu2023Visual,openai2024gpt4technicalreport}.
Despite great success, SSL encoders are vulnerable to backdoor attacks~\cite{li2023Embarrassingly,carlini2021Poisoning,saha2022Backdoor,jia2022BadEncoder,chen2025backdooring}, which severely undermine their reliability and safety on downstream tasks. An attacker can manipulate model representations by injecting malicious samples into unlabeled training data or by modifying model weights~\cite{carlini2021Poisoning,jia2022BadEncoder}. The compromised encoder behaves normally on clean inputs but exhibits malicious behavior on inputs that contain a trigger pattern. Since such encoders often serve as the backbone for a wide range of downstream tasks, developing effective backdoor defenses is essential for artificial intelligence systems in the era of large-scale pretraining~\cite{liu2023Visual}.

\begin{figure}[t]
    \centering
    \includegraphics[width=\linewidth]{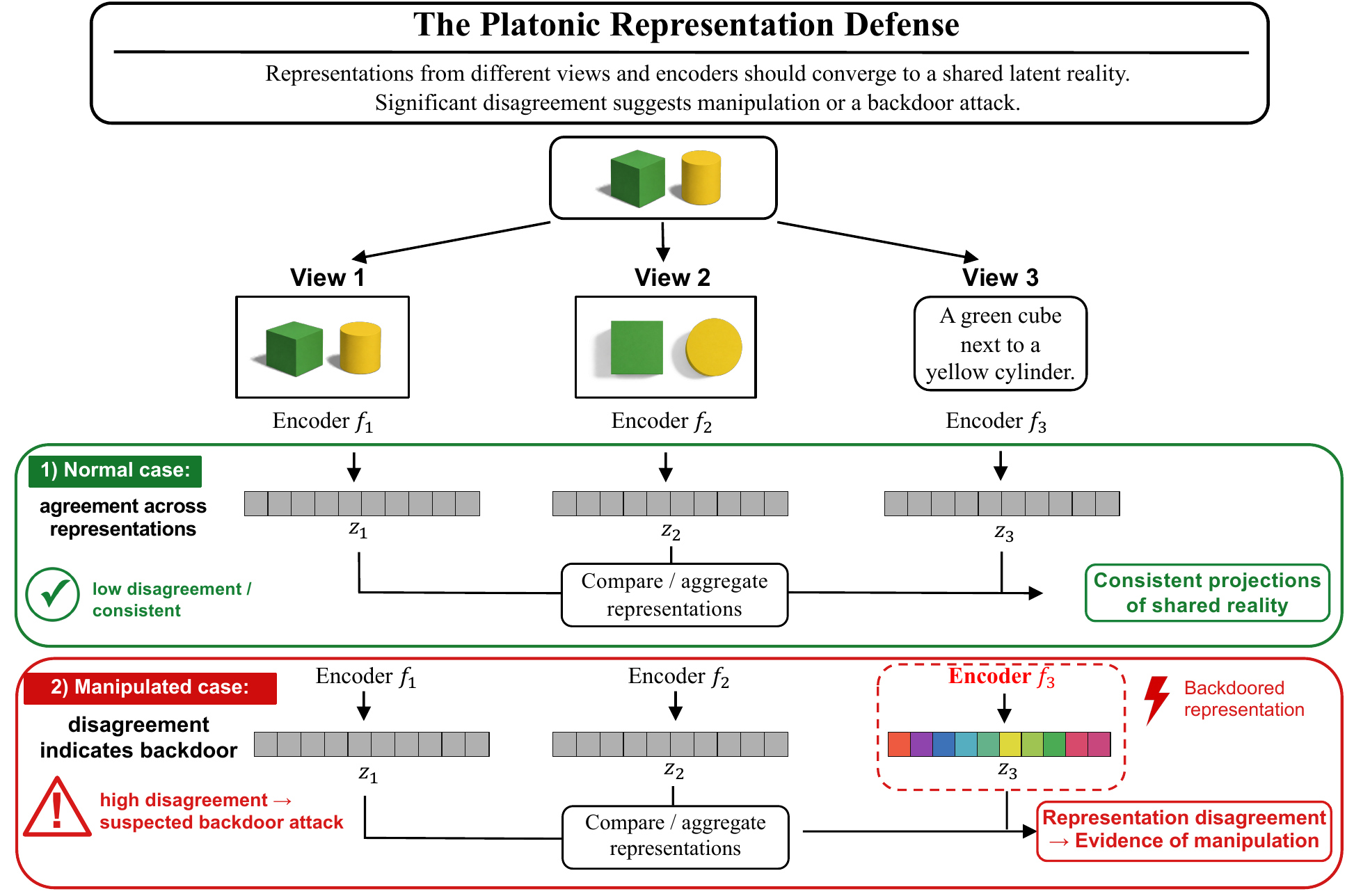}
    \caption{\textbf{The Platonic Representation Defense:}
    Representations ($z_1, z_2, z_3$) are projections of a shared underlying reality. We conjecture that backdoor attacks
    will deviate from this pattern.}
    \label{fig:idea}
    \vspace{-0.5cm}
\end{figure}

Although many defense strategies have been proposed to mitigate backdoor attacks on SSL encoders, only a few operate in a fully black-box setting. Some studies construct pseudo-labels and reformulate the problem under the framework of supervised backdoor defense~\cite{tejankar2023Defending,huang2025DBSSL}, which only applies to a closed and finite label space. Other methods restrict the defense scope to specific types of backdoor triggers~\cite{he2025Closer,tejankar2023Defending} or multimodal backdoor attacks~\cite{bansal2023CleanCLIP}. Their defense capability depends on access to the model's internal structure or the upstream training process. Such knowledge is difficult to obtain in advance in SSL settings, since SSL models serve a broad range of downstream users who typically cannot access the upstream pretraining data or training procedure.

To make defenses attack-agnostic, model-agnostic, and modality-agnostic, we propose a new framework that we call the \emph{Platonic Representation Defense}. The name is borrowed from the Platonic Representation Hypothesis~\cite{huh2024Position}, which hypothesizes that representations produced by different encoders are projections of one common underlying reality and that large-scale pre-training drives these projections toward a shared limit. The compatibility between representations can thus be a good indicator of backdoor attacks if the hypothesis is true. We illustrate this intuition in Figure~\ref{fig:idea}.
We formalize this intuition as a conditional energy function $E_\theta\bigl(z^{\mathsf{s}},\, z^{\mathsf{r}_{1:K}}\bigr)$, where $z^{\mathsf{s}}$ is the output of a source encoder under inspection and $z^{\mathsf{r}_{1:K}}$ are the outputs of $K$ trusted reference encoders on the same input. The energy is low on clean (compatible) samples and rises sharply on backdoored (incompatible) ones, recasting backdoor defense as a problem solved entirely in representation space, without any downstream labels and without access to poisoned samples. We train this energy with Noise Contrastive Estimation (NCE), which sidesteps the intractable partition function while delivering a calibrated detection score. To additionally support purification, we train a denoiser of the source representation conditioned on the reference representations via denoising score matching, by which Tweedie's identity recovers the score field of the same energy. Both objectives share a single architecture and only diverge at the output head and the loss.
Theoretically, we show that when $E_\theta$ is close to the Bayes optimal, the energy gap between the matched (clean) and the mismatched (backdoor) distributions is lower bounded by the mutual information $I\bigl(Z^{\mathsf{s}};\, Z^{\mathsf{r}_{1:K}}\bigr)$ between the source and reference representations. Under the Platonic Representation Hypothesis, independently trained encoders retain shared information, which gives clean representations a systematically lower energy than mismatched ones.

Our main contributions can be summarized as follows.
\textbf{(1)}~To our knowledge, this is the first work to connect representation convergence with backdoor defense, identifying the convergence of independently trained encoders as a previously unused security signal that is intrinsic to large-scale pre-training.
\textbf{(2)}~We introduce the platonic representation defense, a black-box test-time framework that formalizes multi-encoder compatibility as a conditional energy model and realizes it through two complementary training routes, discriminative NCE for detection and conditional denoising score matching for purification.
\textbf{(3)}~We provide both theoretical and empirical evidence. We validate our method on diverse SSL encoders against more than 10 backdoor attacks. Our framework preserves clean sample performance well, which makes it better suited to black-box settings.

\section{Preliminaries}
\label{sec:preliminaries}

We focus on the visual representations, leaving the diagnosis of representations in other modalities for future work.
A visual representation is defined as a function $f: \mathcal{X} \to \mathcal{Z}$, which maps an input image $x \in \mathcal{X}$ to a latent vector $f(x) \in \mathcal{Z}$.
We characterize visual representations by their induced similarity structure, referred to as the kernel.
Evaluating visual representations via kernels is a widely adopted approach~\cite{huh2024Position,wang2020Understanding,kornblith2019similarity}, which can be justified by the fact that the similarity structure between two representations is invariant to the selection of feature space.
In our experiments, we mainly use the kernel-alignment metric Center-Kernel Nearest Neighbor Agreement (CKNNA)~\cite{huh2024Position} to quantify the similarity between two representations.

\paragraph{Representation convergence.}
A growing body of work shows that, as model capacity and training data scale up, independently trained models converge toward increasingly similar kernels.
The \emph{Plato Representation Hypothesis}~\cite{huh2024Position,koepke2026Back} formalizes this observation: the kernel-alignment between two models grows monotonically with their capability, irrespective of architecture, training objective, or even input modality, suggesting that sufficiently capable models recover a shared statistical structure underlying visual data.
Earlier studies of cross-architecture and layer-wise similarity~\cite{kornblith2019similarity,raghu2021do,nguyen2021do} report the same trend at smaller scales, and recent evidence further suggests that the shared kernel reflects intrinsic structure of the visual world rather than arbitrary learned conventions~\cite{cai2025Computer,acevedo2025quantitative}.

\paragraph{Energy-based models and EDM parameterization.}
An energy-based model (EBM)~\cite{lecun2006tutorial} parameterizes a probability density through a scalar energy function $E_\theta : \mathcal{Z} \to \mathbb{R}$ as
\begin{equation}
    p_\theta(z) \;=\; \frac{1}{Z(\theta)}\exp\!\bigl(-E_\theta(z)\bigr),
    \qquad
    Z(\theta) \;=\; \int \exp\!\bigl(-E_\theta(z)\bigr)\,\mathrm{d}z,
    \label{eq:ebm_density}
\end{equation}
trading the normalization constraint of standard likelihood models for a free-form energy landscape. 
The partition function $Z(\theta)$ is intractable, so we bypass it with noise-contrastive estimation (NCE)~\cite{gutmann2012noise}
and score matching~\cite{hyvarinen2005estimation} which fits score $-\nabla_z E_\theta(z)$ without evaluating $Z(\theta)$.
We adopt the practical EDM~\cite{karras2022elucidating} variant of score matching used by modern diffusion models~\cite{song2023Consistency,luo2023latentconsistencymodelssynthesizing}.
For a clean data point $z \sim p_{\mathrm{data}}$, a Gaussian perturbation $\epsilon \sim \mathcal{N}(0, I)$, and a noise level $\sigma > 0$, define
\begin{equation}
    \tilde z \;=\; z + \sigma\,\epsilon,
    \qquad
    p_\sigma(\tilde z) \;=\; \int \mathcal{N}(\tilde z; z, \sigma^2 I)\,p_{\mathrm{data}}(z)\,\mathrm{d}z.
    \label{eq:edm_forward}
\end{equation}
Rather than regressing on $\nabla_{\tilde z} \log p_\sigma$ directly, score matching learns a \emph{denoiser} $\hat D_\theta(\tilde z, \sigma) \in \mathcal{Z}$ that predicts the clean $z$ from $\tilde z$, trained by the EDM-weighted regression
\begin{equation}
    \mathcal{L}_{\mathrm{DSM}}(\theta)
    \;=\;
    \mathbb{E}_{\sigma,\, z,\, \epsilon}\!\Bigl[\,\lambda(\sigma)\,\bigl\lVert \hat D_\theta\bigl(z + \sigma\epsilon,\, \sigma\bigr) - z \bigr\rVert_2^{2}\,\Bigr].
    \label{eq:dsm_loss_generic}
\end{equation}
Its optimum is the Bayes-optimal denoiser $\hat D^{\star}(\tilde z, \sigma) = \mathbb{E}[z \mid \tilde z, \sigma]$, and Tweedie's identity ties it back to the score:
\begin{equation}
    \nabla_{\tilde z} \log p_\sigma(\tilde z)
    \;=\;
    \bigl(\hat D^{\star}(\tilde z, \sigma) - \tilde z\bigr)\big/\sigma^{2},
    \label{eq:tweedie}
\end{equation}
so any denoiser trained with Eq.~\eqref{eq:dsm_loss_generic} directly induces a score estimator $s_\theta(\tilde z, \sigma) = (\hat D_\theta(\tilde z, \sigma) - \tilde z)/\sigma^{2}$ that recovers $-\nabla_{\tilde z} E_\theta$ on the Gaussian-smoothed EBM.
Following the EDM parameterization~\cite{karras2022elucidating}, we do not have the network predict $z$ from $\tilde z$ directly; instead we expose a residual subnetwork $F_\theta$ and apply fixed $\sigma$-dependent preconditioning,
\begin{equation}
    \hat D_\theta(\tilde z, \sigma)
    \;=\;
    c_{\mathrm{skip}}(\sigma)\,\tilde z \;+\; c_{\mathrm{out}}(\sigma)\,F_\theta\!\bigl(c_{\mathrm{in}}(\sigma)\,\tilde z,\; c_{\mathrm{noise}}(\sigma)\bigr),
    \label{eq:edm_precond}
\end{equation}
with coefficients chosen so that both the input and the regression target of $F_\theta$ have unit variance at every $\sigma$~\cite{karras2022elucidating}:
\begin{equation}
    c_{\mathrm{skip}}(\sigma) = \tfrac{\sigma_{\mathrm{data}}^2}{\sigma^2 + \sigma_{\mathrm{data}}^2},\quad
    c_{\mathrm{out}}(\sigma) = \tfrac{\sigma\,\sigma_{\mathrm{data}}}{\sqrt{\sigma^2 + \sigma_{\mathrm{data}}^2}},\quad
    c_{\mathrm{in}}(\sigma) = \tfrac{1}{\sqrt{\sigma^2 + \sigma_{\mathrm{data}}^2}},\quad
    \lambda(\sigma) = \tfrac{\sigma^2 + \sigma_{\mathrm{data}}^2}{\sigma^2\,\sigma_{\mathrm{data}}^2},
    \label{eq:edm_coeffs}
\end{equation}
and $c_{\mathrm{noise}}(\sigma) = \tfrac14 \log \sigma$, supplied to $F_\theta$ as a sinusoidal time embedding~\cite{vaswani2017attention} so that the noise scale enters the network as an additive token-level signal. Here $\sigma_{\mathrm{data}}$ is the empirical standard deviation of $z$ after a per-dimension standardization step that we apply to every encoder output before training.

\begin{figure*}[t]
  \centering
  \begin{minipage}[t]{0.45\textwidth}
    \centering
    \includegraphics[width=\linewidth]{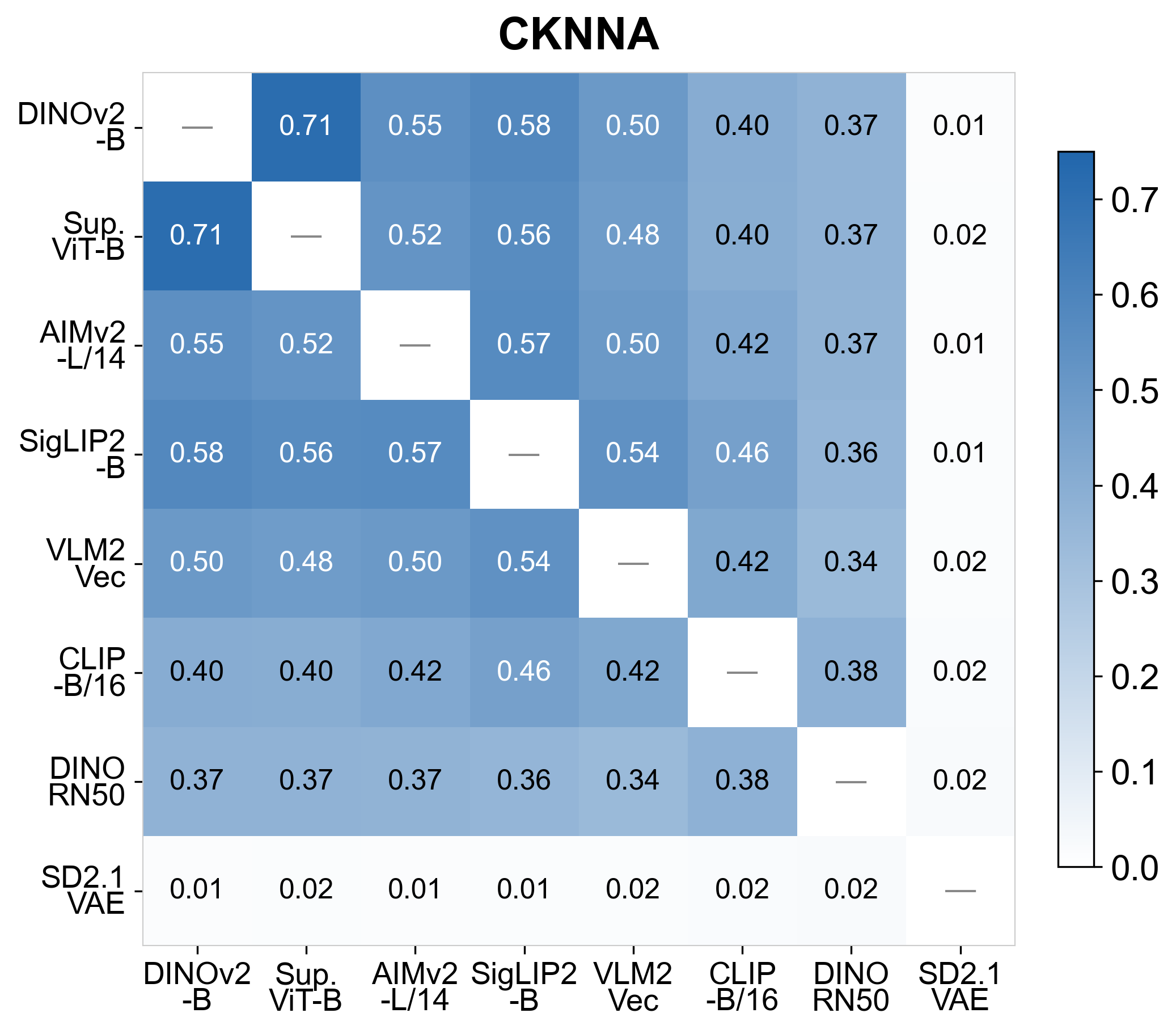}
  \end{minipage}\hfill
  \begin{minipage}[t]{0.51\textwidth}
    \centering
    \includegraphics[width=\linewidth]{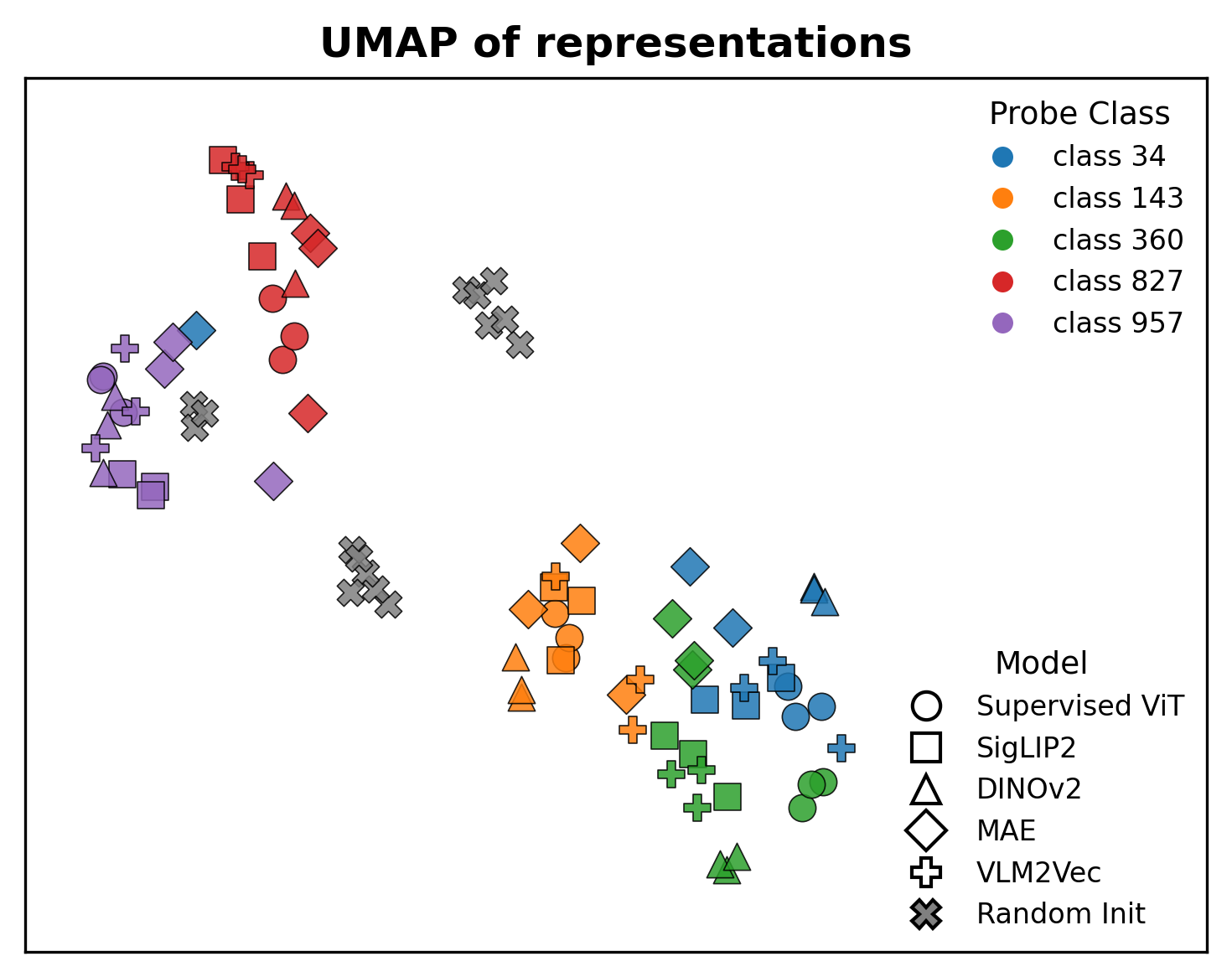}
  \end{minipage}
  \caption{\textbf{Cross-model representation alignment on ImageNet-1K}.
  \textbf{LEFT:}~Pairwise CKNNA scores with 10 neighbors among 8 diverse independently trained models. All discriminative models exhibit substantial mutual alignment despite differing architectures and training objectives. SD-VAE~\cite{rombach2021highresolution} serves as comparision.
  \textbf{RIGHT:}~UMAP of the jointly embedded relational representations after per-model centering and row-$\ell_2$ normalization. Samples from different models cluster primarily by semantic class rather than by model identity, providing a geometric view of the convergence.}
  \label{fig:alignment_empirical}
  \vspace{-0.5em}
\end{figure*}

\section{The Platonic Representation Defense}
\label{sec:platonic_representation_defense}

In this section, we evaluate the aforementioned phenomenon of representation convergence and propose an energy-based modeling framework to model the compatibility of representations.

\paragraph{Threat model.}

The attacker controls upstream pre-training and releases a backdoored encoder $\tilde f$ that behaves normally on clean inputs ($\tilde f(x) \approx f(x)$) but, on any input carrying a predefined trigger, maps the image into the embedding cluster of an attacker-chosen target class~\cite{jia2022BadEncoder,carlini2021Poisoning,saha2022Backdoor,liang2024BadCLIP}.
We consider diverse backdoor families in existing literature including additive image-space patches~\cite{saha2022Backdoor,carlini2021Poisoning}, optimization backdoor~\cite{sun2023Backdoor}, and frequency-domain backdoor~\cite{li2023Embarrassingly}.
The defender accesses neither the upstream pipeline nor the training data. The defender can access publicly available encoders such as DINOv2~\cite{oquab2024dinov2} and CLIP~\cite{radford2021Learning}. The goal is to build test-time detectors and purifiers for poisoned representations without degrading clean-input utility.
We additionally evaluate on adversarial attacks~\cite{goodfellow2014explaining,madry2018towards} to further demonstrate the effectiveness of our method.

\paragraph{Empirical observation.}
We investigate four families of models that span different training paradigms and modalities:
\textbf{(i)}~\emph{Visual SSL encoders} (e.g., DINOv2~\cite{oquab2024dinov2}): self-supervised models that produce semantically rich representations without label supervision;
\textbf{(ii)}~\emph{Vision--language encoders} (e.g., CLIP~\cite{radford2021Learning}): contrastive models that align visual and textual modalities in a shared embedding space;
\textbf{(iii)}~\emph{Multi-modal Large Language Models (MLLMs)} (e.g., Qwen2-VL~\cite{wang2024qwen2}): Large language models with visual encoders;
\textbf{(iv)}~\emph{Generative Encoders} (e.g., AIMv2~\cite{fini2025Multimodal}): Generative encoders trained with autoregressive pre-training~\cite{fini2025Multimodal} or MAE-style pre-training~\cite{he2022Masked}.
We extracted a global representation for Qwen2-VL following VLM2Vec~\cite{jiang2025vlmvec}.
As shown in Figure~\ref{fig:alignment_empirical}, substantial alignment emerges among all models.
Appendix Figure~\ref{fig:alignment_empirical_appendix} shows that randomly initialized counterparts of the same architectures exhibit near-zero alignment ($<$0.01 CKNNA), confirming that alignment arises from learning other than architectural inductive bias.
This convergence has a security implication that, to our knowledge, has not been exploited. \emph{If two independently trained models agree on what an image ``means,'' then disagreement is evidence of manipulation.}

We operationalize this intuition by modeling the joint compatibility of a source representation and a fixed ensemble of $K$ clean reference representations as an energy function.
Given a source encoder $f^{\mathsf{s}}$ (potentially compromised) and $K$ independently trained reference encoders $\{f^{\mathsf{r}_k}\}_{k=1}^{K}$ drawn from the families above, we extract
$z^{\mathsf{s}} = f^{\mathsf{s}}(x) \in \mathbb{R}^{d_s}$ and $z^{\mathsf{r}_k} = f^{\mathsf{r}_k}(x) \in \mathbb{R}^{d_{r_k}}$,
and introduce a scalar \emph{platonic energy}
\begin{equation}
    E_\theta\!\bigl(z^{\mathsf{s}}, z^{\mathsf{r}_{1:K}}\bigr) \in \mathbb{R}
    \label{eq:platonic_energy}
\end{equation}
that assigns each tuple a real-valued compatibility score.
Following the standard energy-based modeling convention~\cite{lecun2006tutorial}, $E_\theta$ induces an unnormalized conditional density over the source latent,
\begin{equation}
    p_\theta\!\bigl(z^{\mathsf{s}} \,\big|\, z^{\mathsf{r}_{1:K}}\bigr) \;\propto\; \exp\!\bigl(- E_\theta(z^{\mathsf{s}}, z^{\mathsf{r}_{1:K}})\bigr),
    \label{eq:conditional_density}
\end{equation}
whose low-energy region captures tuples in which the source and reference representations agree.
The reference representations act as clean, uncorruptible \emph{conditions}.
Once trained, $E_\theta$ provides two complementary primitives for defense.
\textbf{Detection:} for a query image $x$ with observed source latent $z^{\mathsf{s}} = \tilde f^{\mathsf{s}}(x)$ and clean reference features $z^{\mathsf{r}_{1:K}}$, the energy $E_\theta(z^{\mathsf{s}}, z^{\mathsf{r}_{1:K}})$ serves directly as a scalar anomaly score.
\textbf{Purification:} we can follow $-\nabla_{z^{\mathsf{s}}} E_\theta$ to pull an observed latent toward this low-energy region, yielding a purified estimate $\hat z^{\mathsf{s}}$ that is consistent with the references.
In Section~\ref{sec:training_platonic_energy}, we elaborate on learning an $E_\theta$ that supports these two uses.

\paragraph{Detection guarantee from representation convergence.}
Before turning to training, we show that an $E_\theta$ that is close to the Bayes-optimal platonic energy inherits a guarantee \emph{directly} from the representation convergence.
Let $P_{SR}$ denote the distribution of matched tuples $(Z^{\mathsf{s}}, Z^{\mathsf{r}_{1:K}})$ extracted from the same image, and let $P_{\bot} := P_S \otimes P_R$ denote the distribution of cross-sample mismatched tuples in which $Z^{\mathsf{s}}$ is drawn independently of the references.
Define the oracle energy
\begin{equation}
    E^{\star}\!\bigl(z^{\mathsf{s}}, z^{\mathsf{r}_{1:K}}\bigr) \;=\; -\log p\bigl(z^{\mathsf{s}} \,\big|\, z^{\mathsf{r}_{1:K}}\bigr) + \psi\bigl(z^{\mathsf{r}_{1:K}}\bigr),
    \label{eq:oracle_energy}
\end{equation}
which is the Bayes-optimal target of Eq.~\eqref{eq:conditional_density} up to a reference-only shift $\psi$~\cite{lecun2006tutorial}.
Then we characterize the detection behavior of $E^{\star}$ and any $E_\theta$ that approximates it in mean.

\begin{proposition}[Oracle energy gap]
\label{prop:oracle_gap}
Assume $P_{S} \ll P_{S \mid R=r}$ for $P_R$-a.e.\ $r$ and the KL terms below are finite. Then the oracle energy $E^{\star}$ satisfies
\begin{equation}
    \mathbb{E}_{P_{\bot}}\!\bigl[E^{\star}\bigr] \;-\; \mathbb{E}_{P_{SR}}\!\bigl[E^{\star}\bigr]
    \;=\; I\bigl(Z^{\mathsf{s}}; Z^{\mathsf{r}_{1:K}}\bigr) \;+\; \mathbb{E}_{P_R}\!\Bigl[D_{\mathrm{KL}}\bigl(P_S \,\|\, P_{S \mid R}\bigr)\Bigr]
    \;\ge\; I\bigl(Z^{\mathsf{s}}; Z^{\mathsf{r}_{1:K}}\bigr).
    \label{eq:oracle_gap}
\end{equation}
\end{proposition}

\begin{proposition}[High-probability rejection of mismatched sources]
\label{prop:high_prob_rejection}
Let $\gamma := \mathbb{E}_{P_{\bot}}[E_\theta] - \mathbb{E}_{P_{SR}}[E_\theta]$ be the effective energy gap and assume $\gamma > 0$.
Suppose further that $E_\theta - \mathbb{E}_{P}[E_\theta]$ is $\kappa^2$-sub-Gaussian under both $P_{SR}$ and $P_{\bot}$ (which is automatic when $E_\theta$ is bounded within $[a, b]$, with $\kappa^2 = (b-a)^2/4$).
Setting the threshold $\tau = \mathbb{E}_{P_{SR}}[E_\theta] + \gamma/2$,
\begin{equation}
    \Pr_{P_{\bot}}\!\bigl[E_\theta \le \tau\bigr] \;\le\; \exp\!\Bigl(- \tfrac{\gamma^2}{8\kappa^2}\Bigr),
    \qquad
    \Pr_{P_{SR}}\!\bigl[E_\theta \ge \tau\bigr] \;\le\; \exp\!\Bigl(- \tfrac{\gamma^2}{8\kappa^2}\Bigr),
    \label{eq:chernoff_rates}
\end{equation}
and consequently, treating $P_{\bot}$ as the positive class with $E_\theta$ scored so that higher values are more anomalous and evaluating AUROC on independent class-conditional draws $X_{\bot} \sim P_{\bot},\, X_{SR} \sim P_{SR}$, $\mathrm{AUROC}(E_\theta) \ge 1 - 2\exp(-\gamma^2 / 8\kappa^2)$.
\end{proposition}

Proofs can be found in Appendix~\ref{app:detection_theory}.
Proposition~\ref{prop:oracle_gap} identifies the mutual information $I(Z^{\mathsf{s}}; Z^{\mathsf{r}_{1:K}})$ as the energy gap which is exactly the quantity whose strict positivity is the empirical content of the Plato Representation Hypothesis.
Proposition~\ref{prop:high_prob_rejection} upgrades the expectation-level gap to a per-sample guarantee via a standard Chernoff bound for sub-Gaussian variables~\cite{wainwright2019high}.

\section{Training the Platonic Energy}
\label{sec:training_platonic_energy}

We pursue two training routes for the platonic energy of Eq.~\eqref{eq:platonic_energy}, sharing the same architecture but differing in readout and loss: a \emph{discriminative} route that fits $E_\theta$ as a scalar compatibility score with NCE for backdoor detection, and a \emph{score-matching} route that fits a conditional denoiser $\hat D_\theta$ which yields a score field for feature purification.

\paragraph{Shared feature-space backbone architecture.}
The network for both routes shares the same first two stages, after which the readout is route-dependent.
\textbf{(i)}~Per-space projection heads $g^{\mathsf{s}}: \mathbb{R}^{d_s} \to \mathbb{R}^{d_{\mathrm{tok}}}$ and $g^{\mathsf{r}_k}: \mathbb{R}^{d_{r_k}} \to \mathbb{R}^{d_{\mathrm{tok}}}$ ($k=1,\ldots,K$) map heterogeneous-dimensional encoder outputs to a shared token dimension $d_{\mathrm{tok}}$ via three-layer LayerNorm-terminated MLPs.
\textbf{(ii)}~A pre-LayerNorm Transformer~\cite{vaswani2017attention} processes the $(K{+}1)$-token sequence, letting the source token attend to every reference token.
The discriminative route reads $E_\theta$ off the post-transformer source/reference tokens via the cosine head of Eq.~\eqref{eq:pebd_logit}.
The score-matching route instead realizes $\hat D_\theta(\tilde z^{\mathsf{s}}, z^{\mathsf{r}_{1:K}}, \sigma)$ with an additional MLP readout $\pi: \mathbb{R}^{d_{\mathrm{tok}}} \to \mathbb{R}^{d_s}$, which maps the post-transformer source token back to the encoder dimension, and a time embedding module $\eta: \mathbb{R} \rightarrow \mathbb{R}^{d_{\mathrm{tok}}}$ to inject noise-level information.
We refer readers to EDM~\cite{karras2022elucidating} for more details on the denoiser parameterization, and to Section~\ref{sec:experiments_details} for a more detailed architectural description.
Figure~\ref{fig:shared_architecture} visualizes the backbone architecture.
Algorithms~\ref{alg:pebd_readout} and~\ref{alg:dsm_readout} concretely describe the training process of two routes.

\begin{figure}[t]
\centering
\begin{minipage}[t]{0.485\linewidth}
\vspace{0pt}
\begin{algorithm}[H]
\small
\caption{Discriminative training.}
\label{alg:pebd_readout}
\begin{algorithmic}[1]
\Require source $z^{\mathsf{s}}$, references $\{z^{\mathsf{r}_k}\}_{k=1}^{K}$, negative count $M$, position encoding $e_{\mathsf{s}},\ldots,e_{\mathsf{r}_k}$
\State $h^{\mathsf{s}}_0 \gets g^{\mathsf{s}}(z^{\mathsf{s}}) + e_{\mathsf{s}}$ \Comment{project to $d_{\mathrm{tok}}$, add space embed}
\For{$k=1,\ldots,K$}
    \State $h^{\mathsf{r}_k}_0 \gets g^{\mathsf{r}_k}(z^{\mathsf{r}_k}) + e_{\mathsf{r}_k}$
\EndFor
\State $[\tilde h^{\mathsf{s}}_\theta,\, \tilde h^{\mathsf{r}_{1:K}}_\theta] \gets \mathrm{Transformer}([h^{\mathsf{s}}_0,\, h^{\mathsf{r}_{1:K}}_0])$
\State $-E_\theta \gets \alpha \cdot \tfrac{1}{K}\!\sum_{k=1}^{K}\! \cos(\tilde h^{\mathsf{s}}_\theta, \tilde h^{\mathsf{r}_k}_\theta) + \beta$ \Comment{Eq.~\eqref{eq:pebd_logit}}
\State \textbf{Training:} draw $M$ mismatched tuples from $P_{\bot}$, recompute $-E_\theta^{(m)}$ via lines 1--5
\State $\mathcal{L}_{\mathrm{NCE}} \!\gets\! -\!\log \mathrm{sig}(-E_\theta) - \!\!\!\sum_{m=1}^{M}\!\!\!\log(1 \!-\! \mathrm{sig}(-E_\theta^{(m)}))$ \Comment{Eq.~\eqref{eq:pebd_nce_loss}}
\Ensure scalar energy $E_\theta \in \mathbb{R}$
\end{algorithmic}
\end{algorithm}
\end{minipage}\hfill
\begin{minipage}[t]{0.495\linewidth}
\vspace{0pt}
\begin{algorithm}[H]
\small
\caption{cDSM training.}
\label{alg:dsm_readout}
\begin{algorithmic}[1]
\Require standardized source $z^{\mathsf{s}}$, references $\{z^{\mathsf{r}_k}\}_{k=1}^{K}$, noise level $\sigma$, position encoding $e_{\mathsf{s}},\ldots,e_{\mathsf{r}_k}$, time embedding module $\eta$
\State $\epsilon \sim \mathcal{N}(0, I_{d_s})$,\quad $\tilde z^{\mathsf{s}} \gets z^{\mathsf{s}} + \sigma\,\epsilon$ \Comment{Eq.~\eqref{eq:edm_forward}}
\State Compute $c_{\mathrm{in}}(\sigma),\, c_{\mathrm{out}}(\sigma),\, c_{\mathrm{skip}}(\sigma),\, c_{\mathrm{noise}}(\sigma)$ \Comment{Eq.~\eqref{eq:edm_coeffs}}
\State $h^{\mathsf{s}}_0 \gets g^{\mathsf{s}}\!\bigl(c_{\mathrm{in}}(\sigma)\,\tilde z^{\mathsf{s}}\bigr) + \eta\!\bigl(c_{\mathrm{noise}}(\sigma)\bigr) + e_{\mathsf{s}}$
\For{$k=1,\ldots,K$}
    \State $h^{\mathsf{r}_k}_0 \gets g^{\mathsf{r}_k}(z^{\mathsf{r}_k}) + e_{\mathsf{r}_k}$
\EndFor
\State $[\tilde h^{\mathsf{s}}_\theta,\, \cdot\,] \gets \mathrm{Transformer}([h^{\mathsf{s}}_0,\, h^{\mathsf{r}_{1:K}}_0])$
\State $F_\theta \gets \pi\!\bigl(\tilde h^{\mathsf{s}}_\theta\bigr) \in \mathbb{R}^{d_s}$ \Comment{residual readout}
\State $\hat D_\theta \gets c_{\mathrm{skip}}(\sigma)\,\tilde z^{\mathsf{s}} + c_{\mathrm{out}}(\sigma)\,F_\theta$ \Comment{EDM precond}
\State \textbf{Training:} $\mathcal{L}_{\mathrm{cDSM}} \gets \lambda(\sigma)\,\bigl\lVert \hat D_\theta - z^{\mathsf{s}} \bigr\rVert_2^{2}$ \Comment{Eq.~\eqref{eq:cdsm_loss}}
\Ensure denoised $\hat D_\theta \in \mathbb{R}^{d_s}$; score $s_\theta = (\hat D_\theta - \tilde z^{\mathsf{s}})/\sigma^{2}$
\end{algorithmic}
\end{algorithm}
\end{minipage}
\vspace{-0.5em}
\end{figure}

\paragraph{Discriminative training: a sigmoid-NCE method.}
We first introduce a discriminative training route that uses sigmoid-NCE to learn the matched-versus-mismatched density ratio underlying the effective gap $\gamma$ in Proposition~\ref{prop:high_prob_rejection}. We parameterize the discriminative training following SigLIP~\cite{zhai2023sigmoid}:
\begin{equation}
    -E_\theta\!\bigl(z^{\mathsf{s}}, z^{\mathsf{r}_{1:K}}\bigr)
    \;:=\;
    \alpha \cdot \tfrac{1}{K}\!\sum_{k=1}^{K}\!\cos\!\bigl(\tilde h^{\mathsf{s}}_\theta,\, \tilde h^{\mathsf{r}_k}_\theta\bigr) \;+\; \beta,
    \label{eq:pebd_logit}
\end{equation}
where $\tilde h^{\mathsf{s}}_\theta, \tilde h^{\mathsf{r}_k}_\theta \in \mathbb{R}^{d_{\mathrm{tok}}}$ are the post-transformer source and reference token embeddings, and $\alpha = e^{\log\alpha} > 0$, $\beta \in \mathbb{R}$ are two learned scalars (with $\beta$ initialised to $-\log M$ so that the initial log-odds match a uniform prior over the $M{+}1$ candidates).
For each clean tuple $(z^{\mathsf{s}}, z^{\mathsf{r}_{1:K}}) \sim P_{SR}$ extracted from a single image (the positive) we draw $M$ mismatched negatives.
The training loss is the standard binary noise-contrastive objective~\cite{gutmann2012noise,oord2018representation}
\begin{equation}
    \mathcal{L}_{\mathrm{NCE}}(\theta)
    \;=\;
    -\,\mathbb{E}_{P_{SR}}\!\bigl[\log \mathrm{sig}\!\bigl(-E_\theta\bigr)\bigr]
    \;-\; M\,\mathbb{E}_{P_{\bot}}\!\bigl[\log\!\bigl(1 - \mathrm{sig}(-E_\theta)\bigr)\bigr],
    \qquad \mathrm{sig}(t) := \tfrac{1}{1+e^{-t}},
    \label{eq:pebd_nce_loss}
\end{equation}
whose Bayes-optimal logit equals $\log\!\bigl(p(z^{\mathsf{s}}\mid z^{\mathsf{r}_{1:K}}) / (M\,p(z^{\mathsf{s}}))\bigr)$, so $E_\theta$ converges in expectation to $E^{\star}$ of Eq.~\eqref{eq:oracle_energy} up to a sum of a $z^{\mathsf{s}}$-only and a $z^{\mathsf{r}}$-only shift that both cancel in the energy gap of Eq.~\eqref{eq:oracle_gap}~\cite{gutmann2012noise}.
The discriminative variant produces only a scalar for compatibility judgement, but it cannot induce the score field that the feature purification requires, motivating the DSM-based instantiation that follows.

\paragraph{Conditional denoising score matching (cDSM) objective.}
We train the network with the conditional form of Eq.~\eqref{eq:dsm_loss_generic},
\begin{equation}
    \mathcal{L}_{\mathrm{cDSM}}(\theta)
    \;=\;
    \mathbb{E}_{\sigma,\, (z^{\mathsf{s}},\, z^{\mathsf{r}_{1:K}}),\, \epsilon}\!\Bigl[\,\lambda(\sigma)\,\bigl\lVert \hat D_\theta\bigl(z^{\mathsf{s}} + \sigma\epsilon,\, z^{\mathsf{r}_{1:K}},\, \sigma\bigr) - z^{\mathsf{s}} \bigr\rVert_2^{2}\,\Bigr],
    \label{eq:cdsm_loss}
\end{equation}
with $\sigma \sim \mathrm{LogUniform}[\sigma_{\min}, \sigma_{\max}]$, $\epsilon \sim \mathcal{N}(0, I_{d_s})$, and $\lambda(\sigma)$ as in Eq.~\eqref{eq:edm_coeffs}.
At its optimum, Eq.~\eqref{eq:cdsm_loss} recovers the Bayes-optimal \emph{conditional} denoiser $\hat D^{\star}(\tilde z^{\mathsf{s}}, z^{\mathsf{r}_{1:K}}, \sigma) = \mathbb{E}[z^{\mathsf{s}} \mid \tilde z^{\mathsf{s}}, z^{\mathsf{r}_{1:K}}, \sigma]$ of the \emph{clean} joint distribution. By Tweedie's identity (Eq.~\eqref{eq:tweedie}), we can derive the conditional score $(\hat D_\theta - \tilde z^{\mathsf{s}}) / \sigma^{2}$ that approximates $\nabla_{\tilde z^{\mathsf{s}}} \log p_\sigma(\tilde z^{\mathsf{s}} \mid z^{\mathsf{r}_{1:K}})$, i.e., the gradient of the implicit platonic energy with respect to the source latent.
For each mini-batch we draw per-sample Bernoulli masks with drop probability 0.1 to implement classifier-free-guidance~\cite{ho2021classifierfree} (CFG).

\section{Experiments}
\label{sec:exp}

\begin{table*}[t]
  \centering
  \caption{The (CA \%, PA \%, AUC, ASR \%) of defense methods against different backdoor attacks on ImageNet-1K. The adversarial attacks are implemented in an untargeted manner. $\dagger$ denotes our implementation.}
  \label{tab:defense_backdoor_attacks}
  \scriptsize
  \setlength{\tabcolsep}{2.5pt}
  \resizebox{\textwidth}{!}{%
    \begin{tabular}{@{}l|ccc|cccc|cccc|ccc|c|c|cccc@{}}
      \toprule
      Attack & \multicolumn{3}{c|}{\begin{tabular}{@{}c@{}}No Defense\end{tabular}} & \multicolumn{4}{c|}{\begin{tabular}{@{}c@{}}Decomp~\cite{he2025Closer}\\(ICML 2025)\end{tabular}} & \multicolumn{4}{c|}{\begin{tabular}{@{}c@{}}DeDe~\cite{hou2025DeDe}\\(CVPR 2025)\end{tabular}} & \multicolumn{3}{c|}{\begin{tabular}{@{}c@{}}ZIP\cite{shi2023blackbox}\\(NeurIPS 2023)\end{tabular}} & \multicolumn{1}{c|}{\begin{tabular}{@{}c@{}}Beatrix~\cite{ma2023Beatrix}\\(NDSS 2023)\end{tabular}} & \multicolumn{1}{c|}{\begin{tabular}{@{}c@{}}DetectCLIP~\cite{huang2025detecting}\\(ICLR 2025)\end{tabular}} & \multicolumn{4}{c}{\begin{tabular}{@{}c@{}}Platonic Defense\\(Ours)\end{tabular}} \\
      \cmidrule(lr){2-4}\cmidrule(lr){5-8}\cmidrule(lr){9-12}\cmidrule(lr){13-15}\cmidrule(lr){16-16}\cmidrule(lr){17-17}\cmidrule(l){18-21}
      & CA $\uparrow$ & PA $\uparrow$ & ASR $\downarrow$ & CA $\uparrow$ & PA $\uparrow$ & AUC $\uparrow$ & ASR $\downarrow$ & CA$^{\dagger}$ $\uparrow$ & PA$^{\dagger}$ $\uparrow$ & AUC $\uparrow$ & ASR$^{\dagger}$ $\downarrow$ & CA $\uparrow$ & PA $\uparrow$ & ASR $\downarrow$ & AUC $\uparrow$ & AUC $\uparrow$ & CA $\uparrow$ & PA $\uparrow$ & AUC $\uparrow$ & ASR $\downarrow$ \\
      \midrule
      \multicolumn{21}{@{}c@{}}{\cellcolor{gray!15}\scriptsize\emph{Backdoor attacks on unimodal encoders}} \\
      \midrule
      SSL-Backdoor~\cite{saha2022Backdoor} & 76.49 & 0.03 & 99.96 & \underline{76.75} & 0.12 & \textbf{1.00} & 99.97 & 31.1 & \underline{0.2} & 0.65 & \underline{37.1} & 66.73 & 0.07 & 100.0 & 0.96 & 0.55 & \textbf{77.47} & \textbf{74.13} & \underline{0.99} & \textbf{0.10} \\
      CorruptEncoder~\cite{carlini2021Poisoning} & 76.25 & 0.08 & 99.90 & \underline{76.35} & 0.16 & \textbf{1.00} & 99.89 & 30.2 & \underline{0.3} & 0.69 & \underline{46.3} & 66.33 & 0.07 & 99.93 & \underline{0.97} & 0.31 & \textbf{77.31} & \textbf{73.94} & \textbf{1.00} & \textbf{0.30} \\
      CTRL~\cite{li2023Embarrassingly} & 76.59 & 0.03 & 99.96 & \underline{76.93} & 0.14 & \textbf{1.00} & 99.96 & 30.1 & 0.2 & 0.93 & 38.5 & 66.40 & \underline{57.20} & \textbf{0.07} & \textbf{1.00} & 0.49 & \textbf{77.42} & \textbf{64.65} & \underline{0.99} & \underline{1.98} \\
      BLTO~\cite{sun2023Backdoor} & 76.66 & 3.76 & 94.44 & \underline{76.87} & 3.28 & \textbf{0.96} & 95.36 & 29.94 & 1.34 & 0.77 & 34.57 & 64.40 & \textbf{63.93} & \textbf{0.40} & \underline{0.89} & 0.60 & \textbf{77.34} & \underline{55.80} & 0.84 & \underline{3.31} \\
      NA~\cite{chen2025backdooring} & 74.68 & 0.10 &  100.0 & \underline{74.98} & 0.11 & \textbf{1.00} & 100.00 & 31.0 & \underline{0.2} & 0.76 & \textbf{0.6} & 65.73 & 0.07 & 100.0 & \underline{0.95} & 0.00 & \textbf{77.41} & \textbf{49.55} & \underline{0.95} & \underline{18.15} \\
      BadEncoder~\cite{jia2022BadEncoder} & 75.78 & 0.10 & 100.0 & \underline{75.83} & 0.11 & \textbf{1.00} & 100.00 & 28.4 & \underline{0.2} & 0.78 & \underline{28.3} & 66.47 & 0.07 & 100.0 & \textbf{1.00} & 0.05 & \textbf{78.05} & \textbf{74.44} & \underline{0.84} & \textbf{0.16} \\
      DRUPE~\cite{tao2023Distribution} & 72.93 & 0.10 & 100.0  & \underline{73.11} & \underline{0.11} & \underline{0.99} & 100.00 & 26.2 & 0.1 & 0.86 & \underline{51.4} & 63.73 & 0.07 & 100.0 & \textbf{1.00} & 0.14 & \textbf{77.32} & \textbf{72.91} & \underline{0.99} & \textbf{0.03} \\
      Average & 75.63 & 0.60 & 99.18 & \underline{75.83} & 0.58 & \textbf{0.99} & 99.31 & 29.56 & 0.36 & 0.78 & \underline{33.82} & 65.68 & \underline{17.35} & 71.49 & \underline{0.97} & 0.31 & \textbf{77.47} & \textbf{66.49} & 0.94 & \textbf{3.43} \\
      \midrule
      \multicolumn{21}{@{}c@{}}{\cellcolor{gray!15}\scriptsize\emph{Backdoor attacks on image-text encoders}} \\
      \midrule
      CLIP-Backdoor~\cite{carlini2021Poisoning} & 53.25 & 0.24 & 99.51 & \underline{54.60} & \underline{8.97} & \underline{0.95} & 65.33 & 14.8 & 1.4 & 0.69 & \underline{22.1} & 43.47 & 1.07 & 98.06 & 0.76 & \textbf{1.00} & \textbf{63.99} & \textbf{57.41} & \underline{0.95} & \textbf{0.00} \\
      BadCLIP~\cite{liang2024BadCLIP} & 51.01 & 0.55 & 98.08 & \underline{52.08} & 0.40 & 0.48 & 96.24 & 16.4 & 1.0 & 0.68 & 10.7 & 40.47 & \underline{19.80} & \underline{5.54} & 0.41 & \textbf{1.00} & \textbf{64.07} & \textbf{52.19}  & \underline{0.94} & \textbf{0.00} \\
      Average & 52.13 & 0.40 & 98.80 & \underline{53.34} & 4.69 & 0.72 & 80.79 & 15.60 & 1.20 & 0.69 & \underline{16.40} & 41.97 & \underline{10.44} & 51.80 & 0.59 & \textbf{1.00} & \textbf{64.03} & \textbf{54.80} & \underline{0.95} & \textbf{0.00} \\
      \midrule
      \multicolumn{21}{@{}c@{}}{\cellcolor{gray!15}\scriptsize\emph{Adversarial Attacks}} \\
      \midrule
      FGSM~\cite{goodfellow2014explaining} & 78.30 & 2.20 & 97.80 & \underline{78.40} & 2.20 & 0.47 & 97.80 & 48.6 & 2.7 & 0.53 & 97.3 & 67.80 & \underline{43.80} & \underline{56.20} & 0.50 & \textbf{0.75} & \textbf{78.6} & \textbf{58.4} & \underline{0.63} & \textbf{41.60} \\
      PGD~\cite{madry2018towards} & 78.30 & 0.00 & 100.00 & \underline{78.40} & 0.00 & 0.45 & 100.00 & 48.6 & 0.00 & 0.54 & 100.00 & 68.00 & \underline{57.50} & \underline{42.50} & 0.30 & \underline{0.89} & \textbf{78.6} & \textbf{61.8} & \textbf{0.92} & \textbf{38.20} \\
      MI-FGSM~\cite{dong2018boosting} & 78.30 & 0.00 & 100.00 & \underline{78.40} & 0.00 & 0.45 & 100.00 & 48.6 & 0.00 & 0.54 & 100.00 & 67.80 & \underline{47.80} & \underline{52.20} & 0.31 & \underline{0.87} & \textbf{78.6} & \textbf{53.8} & \textbf{0.88}  & \textbf{46.20} \\
      CW-L2~\cite{carlini2017towards} & 78.30 & 0.00 & 100.00 & \underline{78.50} & 2.00 & 0.49 & 98.00 & 43.5 & 18.0 & 0.49 & 82.0 & 68.33 & \underline{67.67} & \underline{32.33} & \underline{0.50} & 0.47 & \textbf{78.6} & \textbf{75.0} & \textbf{0.53}  & \textbf{25.00} \\
      AutoAttack~\cite{croce2020reliable} & 78.30 & 0.00 & 100.00 & \underline{77.40} & 0.00 & 0.46 & 100.00 & 44.6 & 2.6 & 0.55 & 97.4 & 68.00 & \underline{60.33} & \underline{39.67} & 0.34 & \textbf{0.96} & \textbf{78.6} & \textbf{69.4} & \underline{0.83} & \textbf{30.60} \\
      Average & 78.30 & 0.44 & 99.56 & \underline{78.22} & 0.84 & 0.46 & 99.16 & 46.78 & 4.66 & 0.53 & 95.34 & 67.99 & \underline{55.42} & \underline{44.58} & 0.39 & \textbf{0.79} & \textbf{78.60} & \textbf{63.68} & \underline{0.76} & \textbf{36.32} \\
      \bottomrule
    \end{tabular}%
  }
\end{table*}

\begin{figure}[t]
  \centering
  \includegraphics[width=\textwidth]{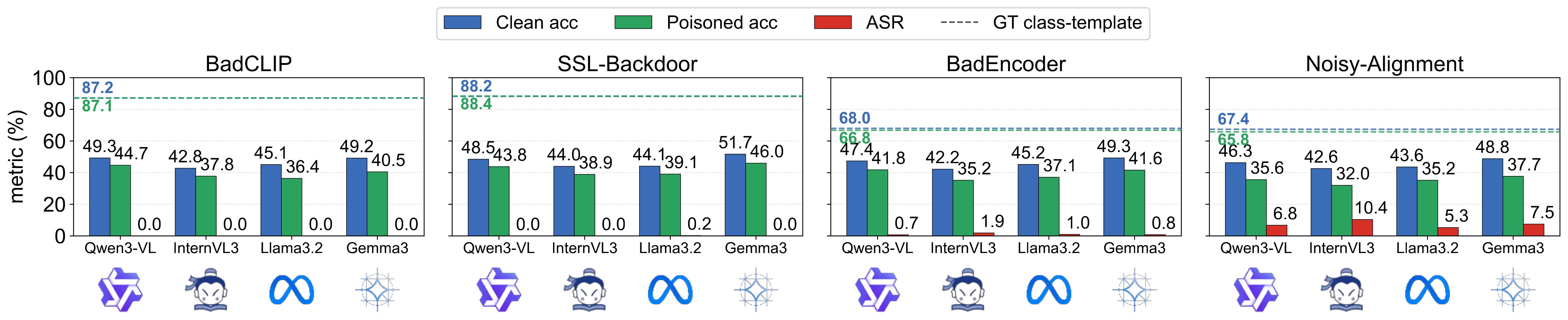}
  \caption{Evaluations on the text condition.}
  \label{fig:textcond_caption_models}
  \vspace{-2em}
\end{figure}

\subsection{Setup}
\label{sec:setup}

For defense evaluation, we consider 3 types of attacks: backdoor attacks to unimodal encoders, backdoor attacks to image-text encoders, and adversarial attacks.
For unimodal visual encoders, we poison DINO ViT-B/16 on ImageNet-1K and cover 7 representative SSL backdoor attacks: SSL-Backdoor~\cite{saha2022Backdoor}, CorruptEncoder~\cite{carlini2021Poisoning}, CTRL~\cite{li2023Embarrassingly}, BLTO~\cite{sun2023Backdoor}, Noisy-Alignment~\cite{chen2025backdooring}, BadEncoder~\cite{jia2022BadEncoder}, and DRUPE~\cite{tao2023Distribution}.
These attacks span patch-based, optimization-based, frequency-domain, and distribution-preserving poisoning mechanisms.
For image--text encoders, we poison CLIP ViT-B/16 on a 60K subset of CC3M and evaluate CLIP-Backdoor~\cite{carlini2021Poisoning} and BadCLIP~\cite{liang2024BadCLIP}.
We additionally test FGSM~\cite{goodfellow2014explaining}, PGD~\cite{madry2018towards}, MI-FGSM~\cite{dong2018boosting}, CW-L2~\cite{carlini2017towards}, and AutoAttack~\cite{croce2020reliable} to assess whether the same representation-compatibility principle extends beyond persistent backdoors.
Detailed settings are provided in Appendix~\ref{sec:experiments_details}.

We compare against recent test-time defenses, including Decomp~\cite{he2025Closer}, DeDe~\cite{hou2025DeDe}, ZIP~\cite{shi2023blackbox}, Beatrix~\cite{ma2023Beatrix}, and DetectCLIP~\cite{huang2025detecting}.
We report clean accuracy (CA), poisoned accuracy (PA), area under the ROC curve (AUC), and attack success rate (ASR). For multimodal encoders, we use zero-shot accuracy~\cite{radford2021Learning} for evaluation. Otherwise, we use linear probing accuracy.
Unless otherwise specified, Platonic Defense uses $K{=}2$ clean reference encoders: the visual branch of SigLIP2-Base/16~\cite{tschannen2025siglip} and DINOv2-Base/16. We use the Heun ODE solver~\cite{ronveaux1995heun} (20 steps) for cDSM inference.
The conditional energy and DSM purifier share the same backbone architecture with 12 layers and hidden dimension 1536, trained only on a subset of ImageNet-1K unavailable during poisoning to avoid overfitting. More details are provided in Appendix~\ref{sec:experiments_details}.

\subsection{Main Results}
\label{sec:exp_main_results}

Table~\ref{tab:defense_backdoor_attacks} shows that platonic defense is the only method that consistently preserves clean utility while suppressing attacks across unimodal backdoors, image--text backdoors, and adversarial perturbations. On 7 unimodal SSL backdoors, it improves the average poisoned accuracy from $0.60\%$ without defense to $66.49\%$ and reduces ASR from $99.18\%$ to $3.43\%$, while slightly increasing clean accuracy to $77.47\%$. Competing defenses either preserve accuracy but leave ASR near $100\%$ (Decomp) or reduce ASR by sacrificing most clean accuracy (DeDe). The same pattern holds for image--text backdoors, where our method reaches $64.03\%$ clean accuracy, $54.80\%$ poisoned accuracy, and $0.00\%$ ASR on average, and extends to untargeted adversarial attacks with the best average poisoned accuracy ($63.68\%$) and lowest ASR ($36.32\%$). Figure~\ref{fig:textcond_caption_models} further shows that the text-conditioned variant remains effective. We train our method on CC3M with the text branch of SigLIP2-Base/16 and CLIP ViT-B/16. Since ImageNet-1K does not provide textual descriptions, we generate image captions using two strategies: (i) class-template labels following CLIP~\cite{radford2021Learning}; and (ii) off-the-shelf VLMs, including Llama3.2-11B-Vision~\cite{grattafiori2024llama3herdmodels}, Gemma3-27B~\cite{gemmateam2025gemma3technicalreport}, Qwen3-VL-4B-Instruct~\cite{bai2025qwen3}, and InternVL3-2B~\cite{zhu2025internvl3}. For all VLMs, we use the prompt: ``Describe this image in one short sentence.''

\begin{figure}[t]
  \centering
  \includegraphics[width=\textwidth]{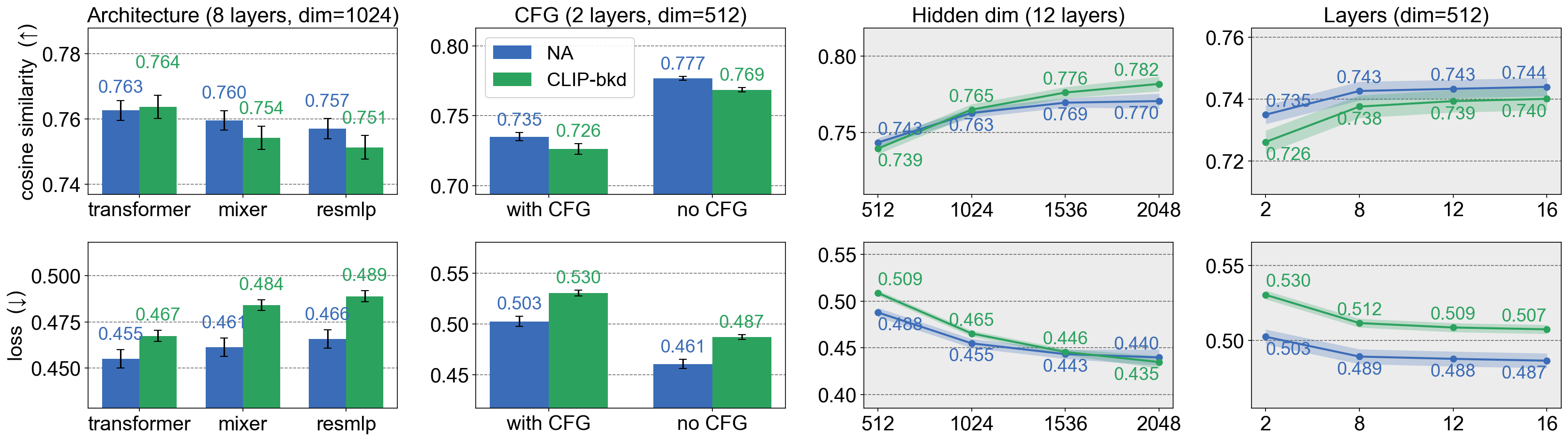}
  \vspace{-2em}
  \caption{Evaluations on the architecture.}
  \label{fig:ablation_architecture}
  \vspace{-1em}
\end{figure}

\begin{table*}[t]
  \centering
  \caption{Defense against enhanced attack. CR means recovered cosine similarity.
  $\checkmark${} denotes a reference that receives the backdoored image.}
  \label{tab:ref_mix_ablation}
  \scriptsize
  \setlength{\tabcolsep}{1.6pt}
  \begin{minipage}[t]{0.49\textwidth}
    \centering
    \begin{tabular}{@{}cccc|cccc@{}}
      \toprule
      DINOv2 & SigLIP2 & AIMv2 & VLM2Vec & CA\,$\uparrow$ & PA\,$\uparrow$ & ASR\,$\downarrow$ & CR\,$\uparrow$ \\
      \midrule
      \multicolumn{8}{@{}c@{}}{\cellcolor{gray!15}\scriptsize\emph{Defense upper bound}} \\
      & & & & \textbf{62.5} & \textbf{62.5} & \textbf{0.0} & \textbf{0.683} \\
      \midrule
      \multicolumn{8}{@{}c@{}}{\cellcolor{gray!15}\scriptsize\emph{One backdoored reference}} \\
      $\checkmark$ & & & & \underline{59.0} & \underline{58.9} & \textbf{0.0} & \textbf{0.654} \\
      & $\checkmark$ & & & 32.7 & 32.4 & 25.7 & 0.460 \\
      & & $\checkmark$ & & 59.6 & 59.6 & \textbf{0.0} & 0.594 \\
      & & & $\checkmark$ & \textbf{60.0} & \textbf{59.9} & \textbf{0.0} & \underline{0.629} \\
      \midrule
      \multicolumn{8}{@{}c@{}}{\cellcolor{gray!15}\scriptsize\emph{Two backdoored references}} \\
      $\checkmark$ & $\checkmark$ & & &  4.3 &  4.1 & 70.3 & 0.374 \\
      $\checkmark$ & & $\checkmark$ & & 49.5 & 49.1 & \textbf{0.0} & 0.532 \\
      $\checkmark$ & & & $\checkmark$ & \textbf{52.7} & \textbf{52.8} & \underline{0.5} & \textbf{0.581} \\
      \bottomrule
    \end{tabular}
  \end{minipage}\hfill
  \begin{minipage}[t]{0.49\textwidth}
    \centering
    \begin{tabular}{@{}cccc|cccc@{}}
      \toprule
      DINOv2 & SigLIP2 & AIMv2 & VLM2Vec & CA\,$\uparrow$ & PA\,$\uparrow$ & ASR\,$\downarrow$ & CR\,$\uparrow$ \\
      \midrule
      \multicolumn{8}{@{}c@{}}{\cellcolor{gray!15}\scriptsize\emph{Two backdoored references}} \\
      & $\checkmark$ & $\checkmark$ & &  8.0 &  7.3 & 46.2 & 0.314 \\
      & $\checkmark$ & & $\checkmark$ &  6.1 &  5.7 & 78.2 & 0.365 \\
      & & $\checkmark$ & $\checkmark$ & \underline{52.6} & \underline{52.7} & \underline{0.5} & \underline{0.515} \\
      \midrule
      \multicolumn{8}{@{}c@{}}{\cellcolor{gray!15}\scriptsize\emph{Three backdoored references}} \\
      $\checkmark$ & $\checkmark$ & $\checkmark$ & &  0.3 &  0.3 & 84.3 & 0.234 \\
      $\checkmark$ & $\checkmark$ & & $\checkmark$ &  0.3 &  0.3 & 96.0 & 0.272 \\
      $\checkmark$ & & $\checkmark$ & $\checkmark$ & \textbf{34.9} & \textbf{34.3} & \textbf{3.9} & \textbf{0.451} \\
      & $\checkmark$ & $\checkmark$ & $\checkmark$ &  0.4 &  0.4 & 67.9 & 0.217 \\
      \midrule
      \multicolumn{8}{@{}c@{}}{\cellcolor{gray!15}\scriptsize\emph{Worst case}} \\
      $\checkmark$ & $\checkmark$ & $\checkmark$ & $\checkmark$ & 0.0 & 0.0 & 100.0 & 0.138 \\
      \bottomrule
    \end{tabular}
  \end{minipage}
\end{table*}

\subsection{Ablation Studies}
\label{sec:exp_ablation}

\paragraph{Evaluation on enhanced attack.}
We consider a more challenging scenario in Table~\ref{tab:ref_mix_ablation}, where we assume the attacker is aware of our defense and has access to the reference encoders.
With all references clean, the defense reaches $62.5\%$ CA/PA and $0.0\%$ ASR. Corrupting one reference is usually tolerated: three of four choices keep ASR at $0.0\%$ with PA around $59$--$60\%$. With two corrupted references, the defense remains effective when the corrupted set excludes SigLIP2 or contains enough architectural diversity (ASR $0.0$--$0.5\%$, PA up to $52.8\%$). Three or four corrupted references largely exceed the intended threat model.

\begin{wrapfigure}[23]{r}{0.6\textwidth}
  \centering
  \includegraphics[width=\linewidth]{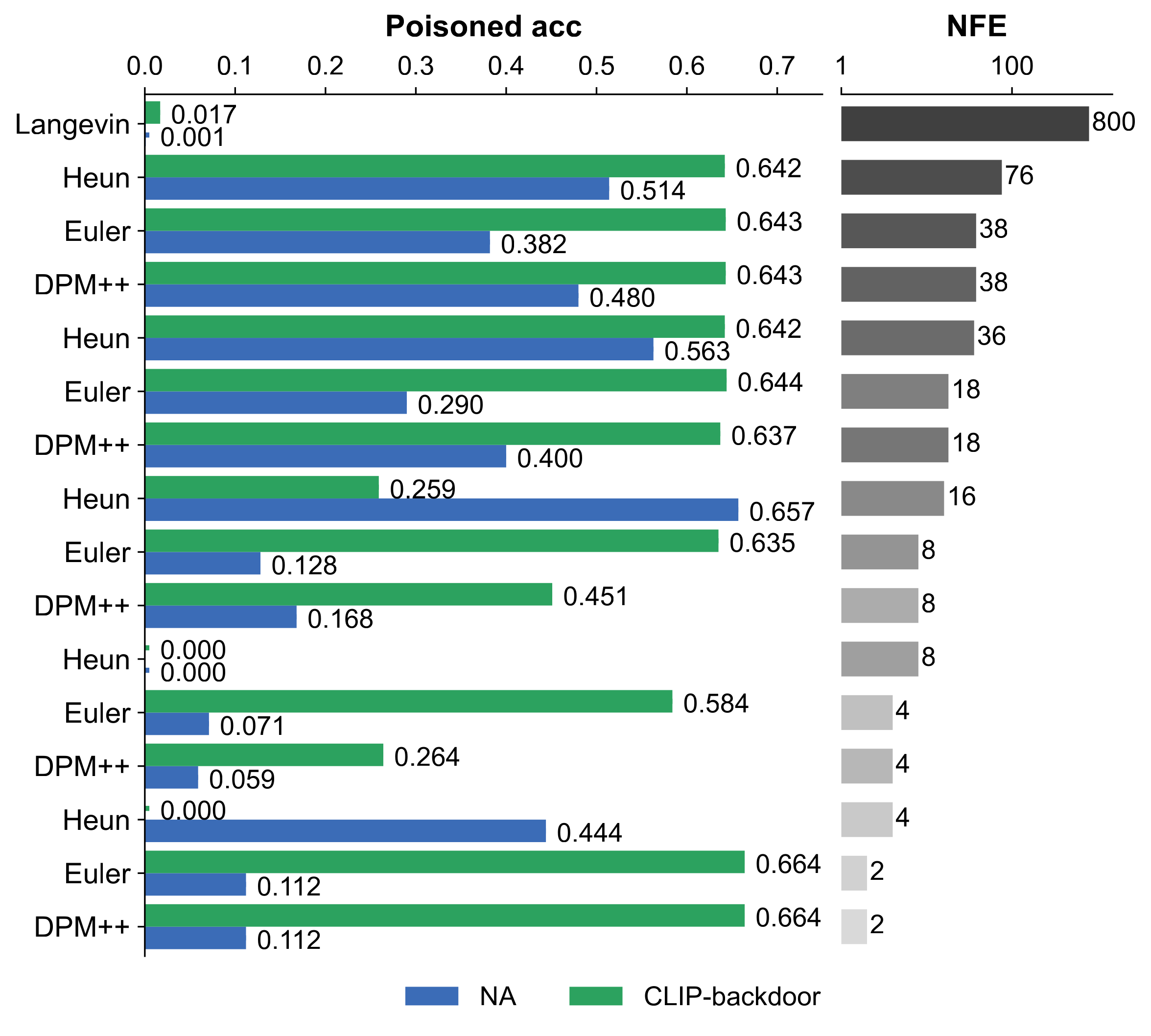}
  \vspace{-2em}
  \caption{Inference efficiency probing.}
  \label{fig:solver_efficiency_simple}
\end{wrapfigure}

\paragraph{Inference overhead.}
Figure~\ref{fig:solver_efficiency_simple} probes the cost--accuracy trade-off of the purification solver. 
Stochastic Langevin sampling~\cite{song2021scorebased} is ineffective even with $800$ network-function evaluations (NFEs). Deterministic ODE solvers are substantially more favorable. The 10-step Heun solver attains the strongest PA in this sweep ($60.3\%$ across the two attacks) with only $36$ NFEs. 
More aggressive settings reduce the cost further, down to $2$--$18$ NFEs for Euler and DPM++~\cite{lu2025dpm}, but their accuracy becomes attack-dependent. 2-step Euler/DPM++ preserves CLIP-Backdoor accuracy while dropping Noisy-Alignment to $11.2\%$. We therefore use 20-step Heun as a conservative default for the main evaluation.

\paragraph{Architecture.}
Figure~\ref{fig:ablation_architecture} ablates the architecture, reporting the denoised cosine similarity and cDSM training loss (eq.~\ref{eq:cdsm_loss}). 
The transformer is consistently the strongest backbone among the tested token mixers: MLP-Mixer~\cite{gui2026adapting} and ResMLP~\cite{li2024return}. Increasing the transformer width has a larger effect than increasing depth: at 12 layers, the cosine improves from $0.743/0.739$ at width 512 to $0.769/0.776$ at width 1536 and saturates near width 2048, whereas increasing depth from 8 to 16 at width 512 changes cosine by less than $0.002$. 
These trends motivate our default 12-layer, 1536-width Transformer, which captures most of the width-scaling gain without paying for the marginal 2048-width improvement. Removing the unconditional branch improves raw denoising metrics, but we keep CFG because it gives a principled way to tune conditioning strength and allows us to drop partial conditions to reduce inference cost when needed.
Table~\ref{tab:share_backbone} tests whether the two objectives can be joint trained. Warm-start
initializes one branch from the model trained for the other objective and then fine-tunes it. Frozen probe directly loads the trained model and evaluates it without adapting the backbone. Joint training optimizes $\mathcal L_{\textsc{nce}}+\mathcal L_{\textsc{cDSM}}$ from scratch.
The transfer is asymmetric. Moving from cDSM to the discriminative branch works well but the reverse direction is less reliable,
especially without fine-tuning. This gap suggests that detection and purification use related but not identical representations.
Joint training largely closes the gap: the discriminative branch stays at baseline AUROC, and
the cDSM branch approaches standalone cDSM performance with only a small loss on
CLIP-Backdoor (\textsc{PA} $0.514$ vs.\ $0.554$) while improving the harder Noisy-Alignment
case (\textsc{PA} $0.596$ vs.\ $0.458$, \textsc{ASR} $0.042$ vs.\ $0.182$).

\begin{table*}[t]
  \centering
  \caption[Backbone transfer between discriminative and cDSM training]{Backbone transfer between the discriminative and cDSM training.
  Gap means platonic energy gap. ``B'' prefix means baseline.}
  \label{tab:share_backbone}
  \scriptsize
  \setlength{\tabcolsep}{3pt}
  \resizebox{\textwidth}{!}{%
    \begin{tabular}{l l | rrrr | rrrr | rrr | rrr}
      \toprule
      & & \multicolumn{4}{c|}{\textsc{B-cdsm}} & \multicolumn{4}{c|}{\textsc{cDSM side}}
        & \multicolumn{3}{c|}{\textsc{B-disc}} & \multicolumn{3}{c}{\textsc{Disc.\ side}} \\
      Source & Variant & Loss & Cos & PA & ASR & Loss & Cos & PA & ASR & Loss & Gap & AUC & Loss & Gap & AUC \\
      \midrule
      \multirow{3}{*}{CLIP-Backdoor}
        & Warm-start   & \multirow{3}{*}{0.571} & \multirow{3}{*}{0.701} & \multirow{3}{*}{0.554} & \multirow{3}{*}{0.000}
                       & 0.605 & 0.678 & 0.496 & 0.000
                       & \multirow{3}{*}{0.144} & \multirow{3}{*}{3.24} & \multirow{3}{*}{0.9999}
                       & 0.147 & 3.18 & 0.9999 \\
        & Frozen probe &  &  &  &  & 0.773 & 0.553 & 0.100 & 0.000 &  &  &  & 0.319 & 1.35 & 0.9954 \\
        & Joint train  &  &  &  &  & 0.591 & 0.688 & 0.514 & 0.000 &  &  &  & 0.143 & 3.23 & 1.0000 \\
      \midrule
      \multirow{3}{*}{Noisy-Alignment}
        & Warm-start   & \multirow{3}{*}{0.581} & \multirow{3}{*}{0.685} & \multirow{3}{*}{0.458} & \multirow{3}{*}{0.182}
                       & 0.619 & 0.659 & 0.466 & 0.116
                       & \multirow{3}{*}{0.143} & \multirow{3}{*}{3.25} & \multirow{3}{*}{0.9988}
                       & 0.148 & 3.17 & 0.9999 \\
        & Frozen probe &  &  &  &  & 0.844 & 0.489 & 0.120 & 0.006 &  &  &  & 0.271 & 1.80 & 0.9972 \\
        & Joint train  &  &  &  &  & 0.603 & 0.671 & 0.596 & 0.042 &  &  &  & 0.142 & 3.26 & 1.0000 \\
      \bottomrule
    \end{tabular}%
  }
\end{table*}

\section{Conclusion}
\label{sec:conclusion}

We introduced a black-box test-time method for defending SSL encoders against backdoor attacks. The
central idea is to use the compatibility between representations from independently trained models as a security signal. 
We formalized it as a conditional energy model and trained it with noise-contrastive estimation and conditional denoising score matching. 
Our analysis connects the energy gap to the mutual information between source and reference representations, giving a
theoretical guarantee for the defense. Empirically, the method improves robustness across multiple SSL encoders and a broad set of attacks, while preserving clean utility. 
Extending the same principle to language, audio, and fully multimodal latent spaces is an important direction for future work.

\clearpage
\bibliographystyle{plainnat}
\bibliography{references}

\appendix
\clearpage
\section{Related Work}
\label{sec:related}

\subsection{Backdoor Attacks on Visual Encoders}
\label{sec:backdoor_attack}

Backdoor attacks on visual encoders pose a severe threat as these encoders serve as general-purpose feature extractors for diverse downstream tasks.
Carlini and Terzis~\cite{carlini2021Poisoning} first demonstrated that contrastive learning is vulnerable to data poisoning based backdoor attacks.
Subsequent works explored different attack vectors: PoisonedEncoder~\cite{liu2022PoisonedEncoder} and Saha~\emph{et al.}~\cite{saha2022Backdoor} poisoned the unlabeled pre-training data, while BadEncoder~\cite{jia2022BadEncoder} directly modified encoder parameters to implant persistent backdoors.
Recent efforts have shifted toward improving attack stealthiness. Tao~\emph{et al.}~\cite{tao2024distribution} designed distribution-preserving attacks, and Zhang~\emph{et al.}~\cite{zhang2025invisible} proposed invisible triggers via frequency-domain manipulation.
With the rise of CLIP, attacks have been extended to multimodal encoders. Liang~\emph{et al.}~\cite{liang2024BadCLIP} leveraged dual-embedding guidance for higher attack success rates.
He~\emph{et al.}~\cite{he2025Closer} systematically analyzed the attack surface of CLIP and revealed new vulnerabilities.

\subsection{Backdoor Detection and Mitigation on Visual Encoders}
\label{sec:backdoor_defense}

Defending visual encoders against backdoor attacks is more challenging than defending supervised classifiers, as SSL encoders lack label information and the backdoor manifests in the representation space.
Feng~\emph{et al.}~\cite{feng2023Detecting} first proposed detecting backdoors in pre-trained encoders by analyzing anomalous representation clusters.
SSL-Cleanse~\cite{zheng2024ssl} combined trigger reverse engineering with mitigation in a unified pipeline.
For post-hoc mitigation, MIMIC~\cite{han2025Mutual} employed mutual information guided knowledge distillation to selectively suppress backdoor features, and Mudjacking~\cite{liu2024mudjacking} repaired compromised neurons via model patching.
Li~\emph{et al.}~\cite{li2024Difficulty} revealed that existing defenses are often insufficient against adaptive attacks on contrastive learning.
For multimodal encoders, CleanCLIP~\cite{bansal2023CleanCLIP} fine-tuned backdoored CLIP with clean contrastive objectives, and Yang~\emph{et al.}~\cite{yang2023robust} incorporated data selection for robust pre-training.
More recently, Huang~\emph{et al.}~\cite{huang2025detecting} detected backdoor samples in CLIP by analyzing cross-modal alignment discrepancies, and DeDe~\cite{hou2025DeDe} leveraged decoder-based reconstruction to identify poisoned inputs for SSL encoders.

\subsection{Similarity on Representations}

Recent studies on representational similarity ask whether independently trained models converge to a common internal organization.
Centered Kernel Alignment (CKA) has become a standard tool for this question, enabling layer-wise comparisons across changes in width, depth, and architecture family~\cite{kornblith2019similarity,nguyen2021do,raghu2021do}.
Using this perspective, Nguyen et al.\ \cite{nguyen2021do} found contiguous blocks of layers whose representations remain strongly aligned despite substantial architectural variation, and showed that probes and layer ablations within these blocks often induce only minor behavioral changes.
This result suggests that overparameterized networks may allocate neighboring layers to closely related computations.
However, high alignment should not be interpreted only as redundancy.
Huh et al.\ \cite{huh2024Position} and Acevedo et al.\ \cite{acevedo2025quantitative} argued that aligned representations can preserve semantically meaningful structure across modalities, including cross-lingual structure within text, while Lad et al.\ \cite{lad2026remarkable} described middle-to-late layers as an incremental feature-construction stage that gradually assembles task-relevant abstractions.
Moving from representational states to representational trajectories, Mahaut and Baroni~\cite{mahaut2026Similarity} showed that layers at similar depths across vision models tend to be closest, but their processing dynamics still differ: classifier models discard low-level image statistics more strongly in later layers, whereas transformer-based models change representations more smoothly than CNNs.
Together, these findings suggest that high-similarity regions can reflect both computational reuse and functionally meaningful stages of representation formation.

Representational similarity has also been studied beyond model-to-model comparisons.
Cai~\emph{et al.}~\cite{cai2025Computer} evaluated 45 foundation and generative models against low-level human visual characteristics, including contrast detection, contrast masking, and contrast constancy.
They found that models trained on vision tasks, especially DINOv2, align with human perceptual thresholds, suggesting that learned feature spaces capture visual structure shared with the human visual system.

\begin{figure*}[t]
  \centering
  \begin{minipage}[t]{0.49\textwidth}
    \centering
    {\small\textbf{(a) CKA}\par}
    \vspace{0.3em}
    \includegraphics[width=\linewidth]{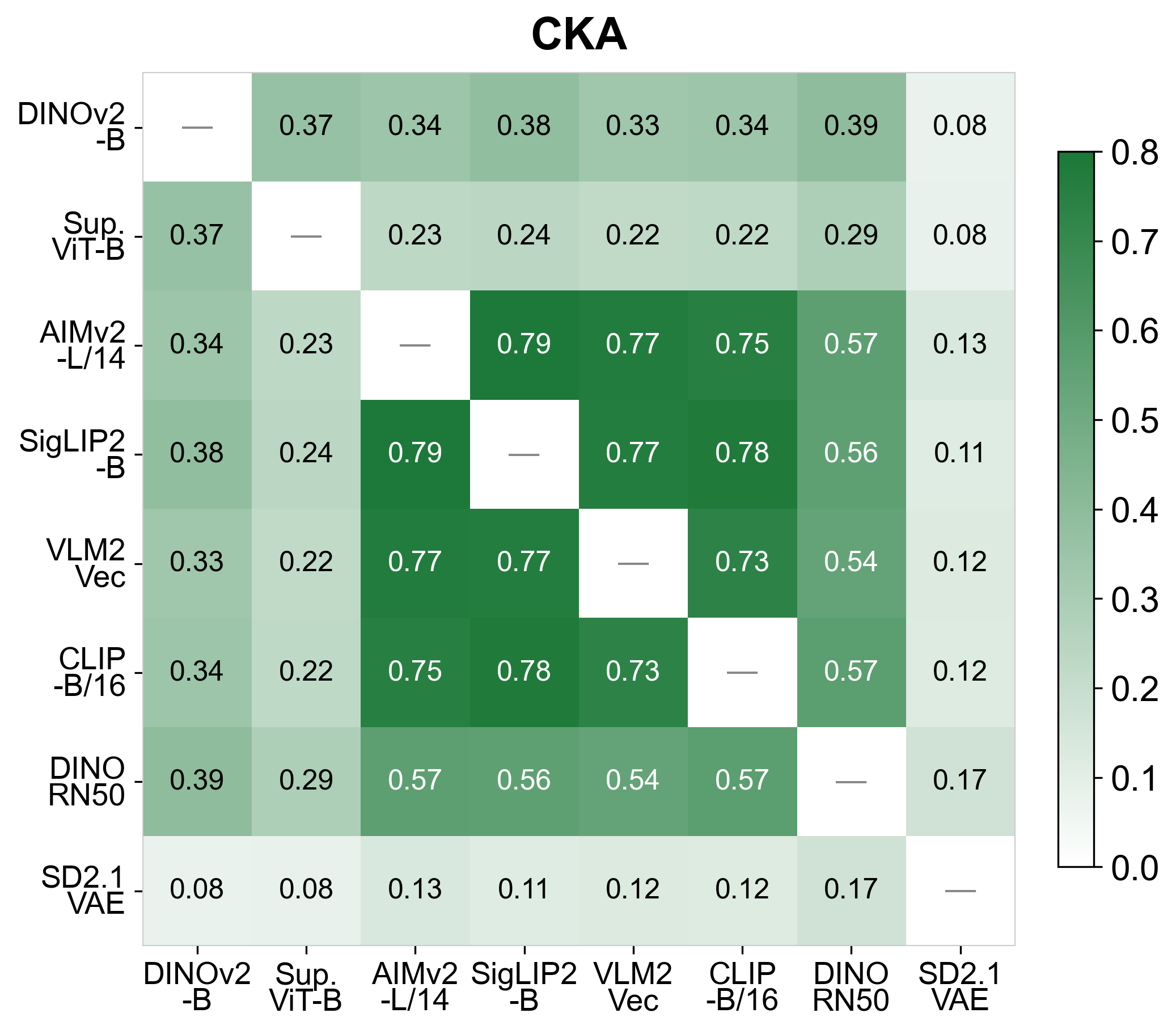}
  \end{minipage}\hfill
  \begin{minipage}[t]{0.49\textwidth}
    \centering
    {\small\textbf{(b) Trained vs.\ random initialization}\par}
    \vspace{0.3em}
    \includegraphics[width=\linewidth]{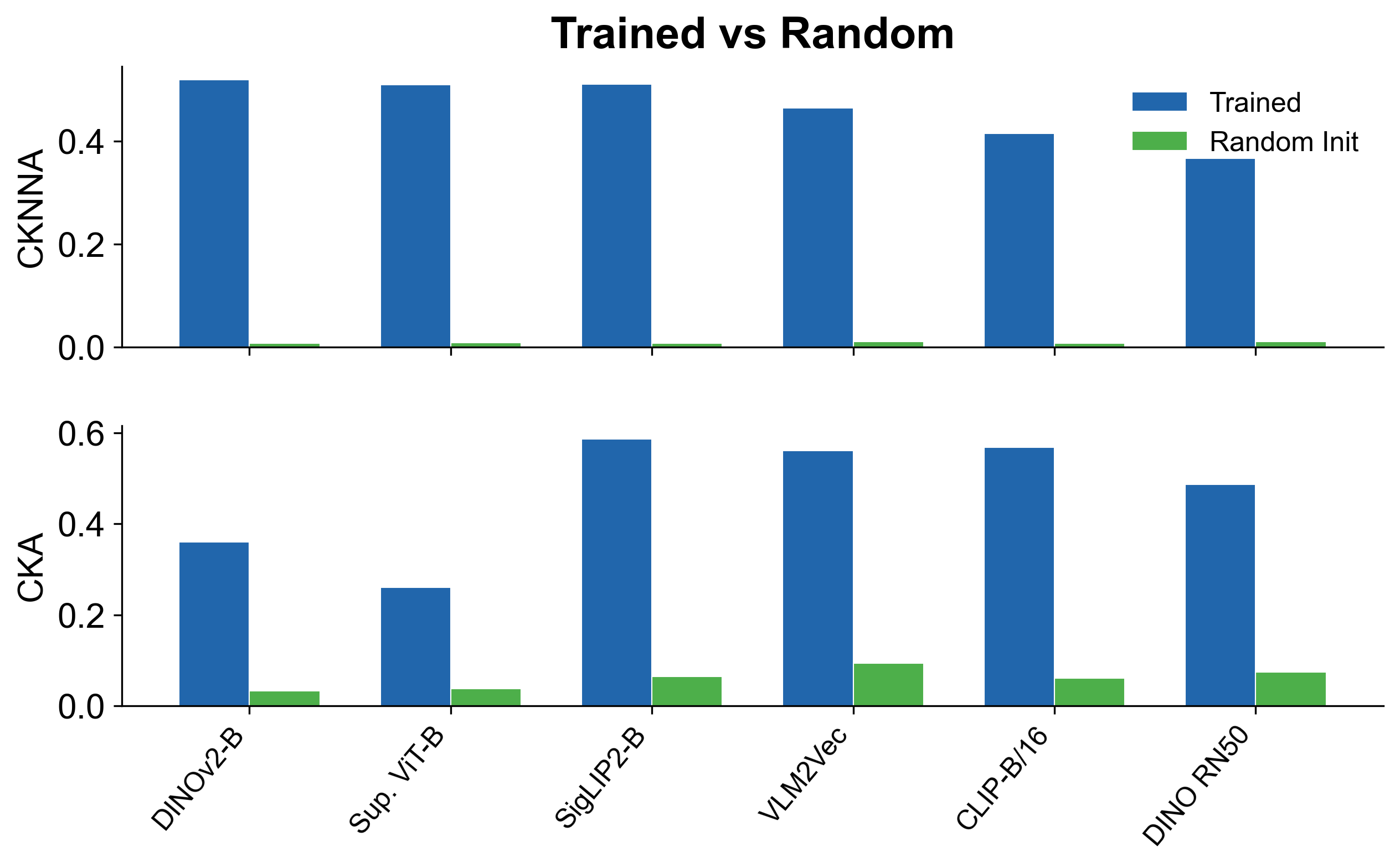}
  \end{minipage}
  \caption{\textbf{Additional alignment visualizations complementing Figure~\ref{fig:alignment_empirical}.}
  \textbf{(a)}~Pairwise CKA scores.
  \textbf{(b)}~Trained models vs.\ their random-initialized counterparts on both metrics: trained models achieve $30$--$100\times$ higher alignment, confirming that the effect arises from learning rather than architectural inductive bias.}
  \label{fig:alignment_empirical_appendix}
\end{figure*}

\section{Limitations}
\label{sec:limitations}

Although the Platonic defense is formulated in a modality-agnostic way, our
empirical study focuses primarily on visual representation spaces. We therefore
leave a systematic evaluation of multimodal conditioning, such as conditioning
vision features on language, audio, or other heterogeneous reference modalities,
to future work. Such settings may introduce a stronger modality mismatch than
the visual-to-visual cases studied in this paper: different modalities can encode
semantic content at different granularities, with different nuisance factors and
latent geometries. As a result, the conditional energy or score model may need
substantially more paired or weakly aligned data to learn a reliable cross-modal
compatibility signal. This requirement could increase both data collection cost
and training cost, especially when multiple reference modalities are used
simultaneously. Understanding how to reduce this data and compute overhead is an
important direction for scaling the method beyond the visual setting.

\section{Proofs for the Detection Guarantee}
\label{app:detection_theory}

Throughout this section, $Z^{\mathsf{s}}$ denotes the source latent and
$Z^{\mathsf{r}}$ denotes the joint reference latent
$Z^{\mathsf{r}_{1:K}}$. Let $P_{SR}$ be their matched joint distribution,
and let $P_{\bot}=P_S\otimes P_R$ be the independent cross-sample product.
We assume that the matched joint distribution admits a regular conditional
density $p(s\mid r)$, and that the source marginal admits a density $p_S(s)$,
so that the conditional cross-entropies and mutual information below are
well defined. We write
$I(S;R):=I(Z^{\mathsf{s}};Z^{\mathsf{r}})$ for brevity.

To avoid degenerate cases in which entropy differences involve undefined
$\infty-\infty$ terms, we make the following standing integrability
assumption:
\begin{equation}
\begin{aligned}
    \mathbb{E}_{P_{SR}}\!\bigl[\,\lvert \log p(S \mid R) \rvert\,\bigr] &< \infty,
    &
    \mathbb{E}_{P_{\bot}}\!\bigl[\,\lvert \log p(S \mid R) \rvert\,\bigr] &< \infty, \\
    I(S; R) &< \infty,
    &
    \mathbb{E}_{P_R}\!\Bigl[D_{\mathrm{KL}}\bigl(P_S \,\|\, P_{S \mid R}\bigr)\Bigr] &< \infty.
\end{aligned}
\label{eq:integrability}
\end{equation}
These regularity conditions are mild for the continuous, normalized latent
representations considered in Section~\ref{sec:exp}; in practice they can be
enforced by the standard small-noise or smoothing view of learned features.

\subsection{Proof of Proposition~\ref{prop:oracle_gap}}
\label{app:proof_prop_oracle_gap}

Let $E^{\star}(s, r) = -\log p(s \mid r) + \psi(r)$ be the oracle energy of Eq.~\eqref{eq:oracle_energy}, with $\psi \in L^1(P_R)$, i.e., $\mathbb{E}_{P_R}[\psi(R)] < \infty$.
Because the $R$-marginals of $P_{SR}$ and $P_{\bot}$ coincide, the reference-only term $\psi(R)$ contributes the same expectation to both populations and cancels from the gap:
\begin{equation}
    \mathbb{E}_{P_{\bot}}\!\bigl[\psi(R)\bigr] = \mathbb{E}_{P_{R}}\!\bigl[\psi(R)\bigr] = \mathbb{E}_{P_{SR}}\!\bigl[\psi(R)\bigr].
    \label{eq:psi_cancels}
\end{equation}
By definition, $\mathbb{E}_{P_{SR}}\!\bigl[-\log p(S \mid R)\bigr] \;=\; H(S \mid R)$. Under $P_{\bot} = P_S \otimes P_R$, we treat $S$ and $R$ as independent and apply Fubini's theorem:
\begin{equation}
    \mathbb{E}_{P_{\bot}}\!\bigl[-\log p(S \mid R)\bigr]
    \;=\; \mathbb{E}_{P_R}\!\biggl[\,\mathbb{E}_{S \sim P_S}\!\bigl[-\log p(S \mid R)\bigr]\biggr].
    \label{eq:mismatched_cross_entropy}
\end{equation}
The inner expectation is the cross-entropy of $P_S$ relative to $P_{S \mid R=r}$.
The cross-entropy decomposes as follows:
\begin{equation}
    \mathbb{E}_{S \sim P_S}\!\bigl[-\log p(S \mid r)\bigr]
    \;=\; H(S) \;+\; D_{\mathrm{KL}}\!\bigl(P_S \,\bigl\|\, P_{S \mid R=r}\bigr),
    \label{eq:cross_entropy_decomposition}
\end{equation}
where $H(S)$ is the differential entropy of $S$ and the KL divergence is non-negative (and, by~\eqref{eq:integrability}, has finite $P_R$-expectation).
Substituting Eq.~\eqref{eq:cross_entropy_decomposition} into Eq.~\eqref{eq:mismatched_cross_entropy} and subtracting $\mathbb{E}_{P_{SR}}\!\bigl[-\log p(S \mid R)\bigr] \;=\; H(S \mid R)$ yields
\begin{align}
    \mathbb{E}_{P_{\bot}}\!\bigl[E^{\star}\bigr] - \mathbb{E}_{P_{SR}}\!\bigl[E^{\star}\bigr]
    &\;=\; \bigl(H(S) - H(S \mid R)\bigr) + \mathbb{E}_{P_R}\!\Bigl[D_{\mathrm{KL}}\bigl(P_S \,\|\, P_{S \mid R}\bigr)\Bigr] \nonumber \\
    &\;=\; I(S; R) + \mathbb{E}_{P_R}\!\Bigl[D_{\mathrm{KL}}\bigl(P_S \,\|\, P_{S \mid R}\bigr)\Bigr] \geq I(S; R).
    \label{eq:oracle_gap_derivation}
\end{align}

\paragraph{Remark on the density-ratio view.}
The same separation can also be viewed through the log-density ratio between
matched and independently paired latents. When the joint and product
distributions have overlapping support, the statistic
\begin{equation}
    A^{\star}(s,r)
    := \log \frac{p_S(s)p_R(r)}{p_{SR}(s,r)}
    = -\log p(s\mid r) + \log p_S(s)
\end{equation}
is the Neyman--Pearson optimal score for distinguishing independent pairs
from matched pairs~\cite{nguyen2010estimating,sugiyama2012density}, with
larger values indicating a more mismatched pair. If the two KL divergences
below are finite, then
\begin{equation}
    \mathbb{E}_{P_{\bot}}[A^{\star}]
    - \mathbb{E}_{P_{SR}}[A^{\star}]
    =
    D_{\mathrm{KL}}(P_{\bot}\,\|\,P_{SR})
    +
    D_{\mathrm{KL}}(P_{SR}\,\|\,P_{\bot}),
    \label{eq:symmetric_kl_gap}
\end{equation}
where the second term is exactly $I(S;R)$. Thus, a detector that estimates
the full matched-vs-independent density ratio also obtains a gap controlled
by the mutual information.

This density-ratio view is mainly useful for discriminative or contrastive
training objectives, such as NCE-style losses. Our oracle argument above only
requires the conditional log-density $-\log p(s\mid r)$, which avoids needing
to model the source marginal term $\log p_S(s)$ explicitly.

\subsection{Proof of Proposition~\ref{prop:high_prob_rejection}}
\label{app:proof_prop_high_prob}

Let $X_{\bot} := E_\theta(Z^{\mathsf{s}}, Z^{\mathsf{r}})$ with $(Z^{\mathsf{s}}, Z^{\mathsf{r}}) \sim P_{\bot}$ and $X_{SR}$ analogously under $P_{SR}$, with means $\mu_{\bot}$, $\mu_{SR}$ and gap $\gamma = \mu_{\bot} - \mu_{SR} > 0$.
By the sub-Gaussian assumption, $X_{\bot} - \mu_{\bot}$ and $X_{SR} - \mu_{SR}$ are $\kappa^2$-sub-Gaussian~\cite[Ch.~2]{wainwright2019high}, so the standard one-sided Chernoff bound yields, for any $t > 0$,
\begin{equation}
    \Pr\!\bigl[X_{SR} - \mu_{SR} \ge t\bigr] \le \exp\!\bigl(-t^2/2\kappa^2\bigr),
    \qquad
    \Pr\!\bigl[\mu_{\bot} - X_{\bot} \ge t\bigr] \le \exp\!\bigl(-t^2/2\kappa^2\bigr).
    \label{eq:two_sided_chernoff}
\end{equation}
Setting the detection threshold $\tau := \mu_{SR} + \gamma/2 = \mu_{\bot} - \gamma/2$ and applying Eq.~\eqref{eq:two_sided_chernoff} with $t = \gamma/2$,
\begin{align}
    \Pr_{P_{SR}}\!\bigl[E_\theta \ge \tau\bigr] &\;=\; \Pr\!\bigl[X_{SR} - \mu_{SR} \ge \gamma/2\bigr] \;\le\; \exp\!\bigl(-\gamma^2 / 8\kappa^2\bigr), \\
    \Pr_{P_{\bot}}\!\bigl[E_\theta \le \tau\bigr] &\;=\; \Pr\!\bigl[\mu_{\bot} - X_{\bot} \ge \gamma/2\bigr] \;\le\; \exp\!\bigl(-\gamma^2 / 8\kappa^2\bigr).
    \label{eq:both_tail_bounds}
\end{align}
By the probability interpretation of the AUROC~\cite{hanley1982meaning}, we have
\begin{equation}
    \mathrm{AUROC}(E_\theta) \;=\; \Pr\!\bigl[X_{\bot} > X_{SR}\bigr] \;+\; \tfrac{1}{2}\Pr\!\bigl[X_{\bot} = X_{SR}\bigr]
    \;\ge\; \Pr\!\bigl[X_{\bot} > X_{SR}\bigr],
\end{equation}
applied to two independent samples $X_{\bot} \sim P_{\bot}, X_{SR} \sim P_{SR}$. Continuity of $E_\theta$ under either marginal (always the case for DSM-trained models with Gaussian noise) merely makes the inequality an equality.

We take a looser \emph{union-bound} version: for independent $X_{\bot}, X_{SR}$,
\begin{equation}
    \{X_{\bot} > \tau\} \cap \{X_{SR} < \tau\} \;\subseteq\; \{X_{\bot} > X_{SR}\},
\end{equation}
so by the union bound on the complement,
\begin{equation}
    \Pr[X_{\bot} > X_{SR}]
    \;\ge\; 1 - \Pr_{P_{\bot}}[E_\theta \le \tau] - \Pr_{P_{SR}}[E_\theta \ge \tau]
    \;\ge\; 1 - 2\exp\!\bigl(-\gamma^2 / 8\kappa^2\bigr),
    \label{eq:auroc_union_bound}
\end{equation}
which yields the main-text bound. 

A tighter rate is available by working directly with the difference variable: writing $Y := X_{\bot} - X_{SR}$ for independent $X_{\bot}, X_{SR}$, $Y$ is sub-Gaussian with parameter $2\kappa^2$ and mean $\gamma$, so $\Pr[Y \le 0] \le \exp(-\gamma^2 / 4\kappa^2)$.
We keep the looser two-tail version in the main text because it also upper-bounds the clean- and mismatched-side false-alarm rates at any single threshold $\tau$.
\hfill$\square$

\section{Experiments Details}
\label{sec:experiments_details}

\begin{figure}[t]
  \centering
  \includegraphics[width=\textwidth]{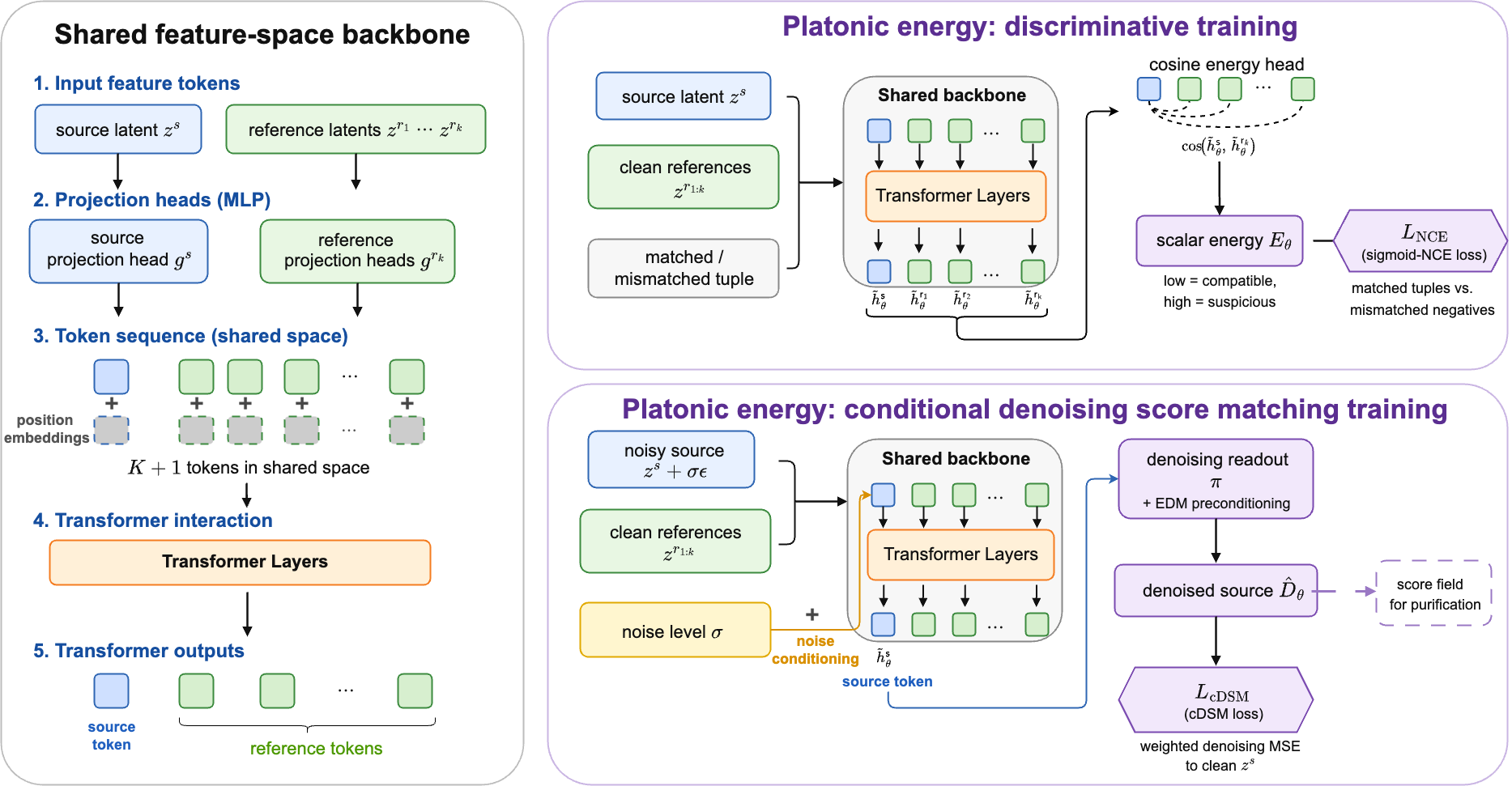}
  \caption{The shared architecture of the platonic energy training for the discriminative and score-matching routes.}
  \label{fig:shared_architecture}
\end{figure}

\paragraph{Architecture instantiation.}
Both training routes share the backbone architecture of Figure~\ref{fig:shared_architecture} and differ only in the readout. Each space is projected to a common token width $d_{\mathrm{tok}}{=}1536$ by a 3-layer MLP, Linear--LayerNorm--GELU--Linear--LayerNorm--GELU--Linear. The discriminative route applies a final LayerNorm to every projected token, while the score-matching route leaves the noised source token unnormalised before EDM preconditioning and applies the final LayerNorm only to reference tokens. A learnable space embedding, initialised from $\mathcal{N}(0,0.02^2)$, is added to each token, and the resulting $1{+}K$ tokens are processed by a 12-layer pre-norm Transformer with 12 attention heads, GELU feed-forward blocks of width $4d_{\mathrm{tok}}$, and zero dropout. We implement the transformer backbone with the encoder-only transformer from Pytorch Library~\cite{paszke2019pytorch}. 
For cDSM, the source token additionally receives a sinusoidal noise embedding with 128 Fourier channels followed by a two-layer SiLU MLP into $d_{\mathrm{tok}}$. The residual denoising head is a Linear--GELU--Linear MLP with hidden width $d_{\mathrm{tok}}$ and a zero-initialised final layer, so the initial denoiser follows the EDM skip connection.

\paragraph{Discriminative route.} The post-transformer source and reference tokens are scored by a cosine readout averaged over the $K$ references; the logit is $\alpha\,\bar{\mathrm{cos}} + \beta$, where $\alpha = \exp(\log\alpha)$ is constrained positive and $\beta$ is initialised to $-\log r$ for a negative ratio $r$ (both learnable). Training uses sigmoid-BCE against negatives produced by a mismatch sampler that mixes \emph{full} (all spaces deranged), \emph{source-only}, and \emph{partial} (a randomly chosen subset of references replaced) deranged tuples; before the feature queue is warm we derange within the mini-batch, afterwards we draw from a per-space FIFO queue.

\paragraph{Score-matching route.} The source feature is first standardised per dimension using statistics $(\mu, \sigma_{\mathrm{dim}})$ pre-computed on clean training data. We adopt the EDM~\citep{karras2022elucidating} preconditioning with $c_{\mathrm{skip}} = \sigma_d^2/(\sigma^2{+}\sigma_d^2)$, $c_{\mathrm{out}} = \sigma\sigma_d/\sqrt{\sigma^2{+}\sigma_d^2}$, $c_{\mathrm{in}} = 1/\sqrt{\sigma^2{+}\sigma_d^2}$, $c_{\mathrm{noise}} = \tfrac14\log\sigma$, where $\sigma_d$ means $\sigma_{\mathrm{data}}$. The source token feeds $c_{\mathrm{in}}\,\tilde z_{\mathrm{std}}$ into the source head and adds a sinusoidal-MLP $\sigma$-embedding alongside the space embedding. A 2-layer MLP readout $F_\theta$ maps the post-transformer source token back to encoder dimension, giving $\hat D_\theta = c_{\mathrm{skip}}\tilde z + c_{\mathrm{out}} F_\theta$ and score $(\hat D_\theta - \tilde z)/\sigma^2$. The DSM loss is the EDM-weighted MSE $\lambda(\sigma)\,\lVert \hat D_\theta - z^{\star}_{\mathrm{std}}\rVert^2$ with $\lambda = (\sigma^2{+}\sigma_d^2)/(\sigma\sigma_d)^2$, $\sigma$ drawn log-uniformly from $[\sigma_{\min},\sigma_{\max}]$, and a $10\%$ probability of zeroing all reference tokens to expose an unconditional branch for classifier-free guidance. At inference we use a Heun-EDM ODE solver over a $\rho{=}7$ schedule by default, with an anchored annealed Langevin sampler available as an alternative. CFG is applied as $s_{\mathrm{uncond}} + w(s_{\mathrm{cond}} - s_{\mathrm{uncond}})$ with $w{=}1$ as the default.

\paragraph{Backdoor implantation.} We instantiate backdoored encoders by post-training pretrained models on small poisoned subsets. For unimodal visual SSL encoders, we poisoned images following corresponding backdoor methods, and optimize the triggered representation with a additional cosine representation preserving loss to prevent catastrophic forgetting.
For image--text encoders, following the CleanCLIP pipeline~\cite{bansal2023CleanCLIP}, we poison 0.6M subset of CC3M by applying the trigger to randomly selected images and replacing their captions with manipulated ImageNet templates, then continue CLIP contrastive training. This procedure makes triggered inputs map to the target semantic region while preserving clean utility.

\begin{table}[ht]
  \centering
  \caption{Training hyperparameters.}
  \label{tab:train_hparams}
  \small
  \begin{tabular}{@{}lll@{}}
    \toprule
    & Discriminative route & Score-matching route \\
    \midrule
    Optimizer & AdamW & AdamW \\
    Peak / min learning rate & $10^{-4}\,/\,10^{-6}$ & $10^{-4}\,/\,10^{-6}$ \\
    Weight decay & $0.01$ & $0.01$ \\
    LR schedule & cosine, $1$ warmup epoch & cosine, $1$ warmup epoch \\
    Gradient clip ($\ell_2$) & $1.0$ & $1.0$ \\
    Effective / micro batch size & $256\,/\,16$ & $256\,/\,16$ \\
    Epochs & $5$ & $5$ \\
    \midrule
    Loss & sigmoid BCE & EDM-weighted DSM \\
    Negative ratio $r$ & $4$ & --- \\
    $\sigma$ sampling & --- & log-uniform $[\sigma_{\min},\sigma_{\max}]$ \\
    Reference-drop prob.\ $p_{\mathrm{drop\_ref}}$ & --- & $0.1$ \\
    EMA decay & --- & $0.999$ \\
    Inference solver / steps & --- & Heun ($\rho{=}7$), $20$ \\
    \bottomrule
  \end{tabular}
\end{table}

\section{Reconstruction-Driven Purification of DEDE}
\label{sec:appendix_dede_purify}

We try to directly optimize the suspicious latent against the reconstruction loss of an off-the-shelf detector, DEDE~\citep{hou2025DeDe}. We evaluated this baseline on
a DINO ViT-B/16 BadEncoder victim~\citep{jia2022BadEncoder} with a
$50{\times}50$ patch trigger. The DEDE decoder is trained
on clean 100K ImageNet-1K data and then frozen. During purification we hold
the encoder and decoder fixed, update only the latent $z=f_\phi(x)$.

The latent update minimizes the masked reconstruction objective
\begin{equation}
  \label{eq:dede_purify_loss}
  \mathcal{L}_{\text{rec}}(x, z) \;=\;
  \mathbb{E}_M\!\left[\frac{1}{|M|}\,\bigl\|\,M\odot \bigl(x - g_\psi(x_{\text{vis}}, z)\bigr)\bigr\|_2^2 \right],
\end{equation}
where $M$ is sampled at each optimizer step. Table~\ref{tab:dede_purify_adam} shows the central result. Stronger Adam updates reduce ASR, but only by simultaneously degrading clean accuracy.
More importantly, the poison Top-1 never exceeds $0.20\%$: even when ASR falls
to zero, the purified latent is not mapped back to the ground-truth class. This
suggests that the reconstruction objective can disrupt the attacker target, but it cannot perceive what the correct representation should be.
This is the failure mode addressed by the conditional DSM purifier, which
learns an explicit score field toward a reference-defined clean representation. We observe the same failure pattern on CLIP-ViT-B/16 victims using a zero-shot CLIP text
classifier. Table~\ref{tab:dede_purify_clip_sweep} shows that CLIP latents are
more sensitive to Adam updates, with usable learning rates around $\eta=0.1$
rather than $\eta=1.0$.

\begin{table}[t]
  \centering
  \caption{\textbf{Latent purification with the \textsc{best-AUROC}
  DEDE decoder on BadEncoder.}
  Evaluation uses a frozen DINO ViT-B/16 BadEncoder, a frozen DEDE decoder, and a deterministic $1{,}000$-image ImageNet-1K
  validation subset.}
  \label{tab:dede_purify_adam}
  \small
  \setlength{\tabcolsep}{6pt}
  \begin{tabular}{@{}cccccc@{}}
    \toprule
    Steps $T$ & LR $\eta$ & Clean $\uparrow$ & Poison $\uparrow$ & ASR $\downarrow$ \\
    \midrule
    $0$ (baseline) & --- & $88.40$ & $0.00$ & $100.00$ \\
    \midrule
    \multirow{3}{*}{$5$}
      & $0.01$ & $88.40$ & $0.00$ & $100.00$ \\
      & $0.10$ & $88.20$ & $0.00$ & $100.00$ \\
      & $1.00$ & $81.70$ & $0.00$ & $100.00$ \\
    \midrule
    \multirow{6}{*}{$20$}
      & $0.01$ & $88.30$ & $0.00$ & $100.00$ \\
      & $0.10$ & $87.80$ & $0.00$ & $100.00$ \\
      & $1.00$ & $67.00$ & $0.00$ &  $79.00$ \\
      & $5.00$ &  $1.50$ & $0.10$ &   $0.10$ \\
      & $20.0$ &  $0.10$ & $0.00$ &   $0.00$ \\
      & $100$  &  $0.00$ & $0.00$ &   $0.00$ \\
    \midrule
    \multirow{6}{*}{$50$}
      & $0.01$ & $88.10$ & $0.00$ & $100.00$ \\
      & $0.10$ & $87.40$ & $0.00$ & $100.00$ \\
      & $1.00$ & $55.20$ & $0.00$ &  $37.70$ \\
      & $5.00$ &  $1.20$ & $0.00$ &   $0.00$ \\
      & $20.0$ &  $0.10$ & $0.00$ &   $0.00$ \\
      & $100$  &  $0.00$ & $0.00$ &   $0.00$ \\
    \midrule
    \multirow{8}{*}{$100$}
      & $0.20$ & $85.50$ & $0.00$ & $100.00$ \\
      & $0.30$ & $82.70$ & $0.00$ & $100.00$ \\
      & $0.50$ & $74.70$ & $0.10$ &  $97.40$ \\
      & $0.70$ & $66.90$ & $0.20$ &  $73.30$ \\
      & $1.00$ & $\mathbf{47.40}$ & $\mathbf{0.10}$ & $\mathbf{22.50}$ \\
      & $5.00$ &  $0.70$ & $0.00$ &   $0.00$ \\
      & $20.0$ &  $0.40$ & $0.00$ &   $0.00$ \\
      & $100$  &  $0.10$ & $0.00$ &   $0.00$ \\
    \bottomrule
  \end{tabular}
\end{table}

\begin{table}[t]
  \centering
  \caption{\textbf{CLIP-ViT-B/16 hyperparameter sweep for DEDE-decoder
  latent purification.}
  Both victims use a frozen CLIP image tower, frozen DEDE decoder, Adam updates, a $1{,}000$-image ImageNet-1k subset, and zero-shot
  CLIP classification.}
  \label{tab:dede_purify_clip_sweep}
  \small
  \setlength{\tabcolsep}{4pt}
  \begin{tabular}{@{}cccccccc@{}}
    \toprule
    & & \multicolumn{3}{c}{\texttt{clip-backdoor}}
    & \multicolumn{3}{c}{\texttt{BadCLIP}} \\
    \cmidrule(lr){3-5}\cmidrule(lr){6-8}
    Steps $T$ & LR $\eta$ & Clean & Poison & ASR & Clean & Poison & ASR \\
    \midrule
    $0$ (baseline) & --- & $50.00$ & $0.30$ & $99.50$ & $47.20$ & $0.60$ & $97.90$ \\
    \midrule
    \multirow{6}{*}{$20$}
      & $0.001$ & $49.50$ & $0.60$ & $99.50$ & $46.60$ & $0.70$ & $97.90$ \\
      & $0.010$ & $48.90$ & $0.30$ & $99.70$ & $46.70$ & $0.70$ & $96.80$ \\
      & $0.050$ & $43.60$ & $0.60$ & $98.00$ & $41.30$ & $1.70$ & $82.48$ \\
      & $0.100$ & $30.00$ & $1.30$ & $54.55$ & $25.90$ & $1.10$ & $30.23$ \\
      & $0.200$ & $\phantom{0}8.60$ & $1.00$ & $\phantom{0}1.60$ & $\phantom{0}6.00$ & $0.40$ & $\phantom{0}2.10$ \\
      & $0.500$ & $\phantom{0}1.60$ & $0.60$ & $\phantom{0}0.00$ & $\phantom{0}1.00$ & $0.10$ & $\phantom{0}0.00$ \\
    \midrule
    \multirow{6}{*}{$50$}
      & $0.001$ & $49.40$ & $0.40$ & $99.40$ & $46.60$ & $0.60$ & $97.90$ \\
      & $0.010$ & $48.30$ & $0.40$ & $99.20$ & $46.10$ & $1.10$ & $95.60$ \\
      & $0.050$ & $38.80$ & $1.10$ & $92.59$ & $38.50$ & $1.60$ & $68.77$ \\
      & $0.100$ & $\mathbf{20.10}$ & $1.20$ & $\mathbf{33.03}$ & $\mathbf{19.00}$ & $0.60$ & $\mathbf{16.92}$ \\
      & $0.200$ & $\phantom{0}5.20$ & $0.40$ & $\phantom{0}1.00$ & $\phantom{0}4.20$ & $0.40$ & $\phantom{0}0.60$ \\
      & $0.500$ & $\phantom{0}0.80$ & $0.30$ & $\phantom{0}0.00$ & $\phantom{0}0.50$ & $0.20$ & $\phantom{0}0.00$ \\
    \midrule
    \multirow{6}{*}{$100$}
      & $0.001$ & $49.30$ & $0.40$ & $99.50$ & $46.50$ & $0.70$ & $97.80$ \\
      & $0.010$ & $47.50$ & $0.40$ & $99.10$ & $44.70$ & $1.10$ & $93.79$ \\
      & $0.050$ & $33.00$ & $1.20$ & $84.18$ & $32.50$ & $1.00$ & $55.26$ \\
      & $0.100$ & $\mathbf{14.80}$ & $1.40$ & $\mathbf{22.12}$ & $\mathbf{16.40}$ & $1.00$ & $\mathbf{10.71}$ \\
      & $0.200$ & $\phantom{0}3.90$ & $0.80$ & $\phantom{0}0.50$ & $\phantom{0}3.40$ & $0.20$ & $\phantom{0}0.40$ \\
      & $0.500$ & $\phantom{0}0.70$ & $0.50$ & $\phantom{0}0.00$ & $\phantom{0}0.60$ & $0.20$ & $\phantom{0}0.20$ \\
    \bottomrule
  \end{tabular}
\end{table}

\end{document}